\documentclass{article}

\usepackage{PRIMEarxiv}

\usepackage[utf8]{inputenc} 
\usepackage[T1]{fontenc}    
\usepackage{hyperref}       
\usepackage{url}            
\usepackage{booktabs}       
\usepackage{amsfonts}       
\usepackage{nicefrac}       
\usepackage{microtype}      
\usepackage{lipsum}
\usepackage{fancyhdr}       
\usepackage{graphicx}       
\graphicspath{{media/}}     

\usepackage{algorithm}
\usepackage{algorithmic}
\usepackage{wrapfig}
\usepackage[utf8]{inputenc} 
\usepackage[T1]{fontenc}    
\usepackage{url}            
\usepackage{booktabs}       
\usepackage{amsfonts}       
\usepackage{nicefrac}       
\usepackage{microtype}      
\usepackage{xcolor}         
\usepackage{amsmath}
\usepackage{amssymb}
\usepackage{mathtools}
\usepackage{amsthm}
\usepackage{mathrsfs}
\usepackage{makecell}
\usepackage{algorithm}
\usepackage{algorithmic}
\usepackage{amssymb}

\usepackage{authblk}
\usepackage{caption}
\usepackage{subcaption}
\theoremstyle{plain}
\newtheorem{theorem}{Theorem}

\newtheorem{lemma}{Lemma}

\theoremstyle{definition}
\newtheorem{definition}{Definition}
\newtheorem{assumption}{Assumption}
\theoremstyle{remark}
\newtheorem{remark}[theorem]{Remark}

\newcommand{\real}{\mathbb{R}}

\newcommand{\policy}{\pi}
\newcommand{\mdp}{\mathcal{M}}
\newcommand{\states}{\mathcal{S}}
\newcommand{\actions}{\mathcal{A}}
\newcommand{\joint}{\mathcal{Z}}

\newcommand{\transitions}{P}

\newcommand{\freq}{d}

\newcommand{\behavior}{{\pi_\beta}}
\newcommand{\bellman}{\mathcal{B}}

\newcommand{\hatbehavior}{\hat{\pi}_\beta}

\newcommand{\bs}{\mathbf{s}}
\newcommand{\ba}{\mathbf{a}}

\newcommand{\E}{\mathbb{E}}
\newcommand{\D}{\mathcal{D}}
\newcommand{\R}{\mathbb{R}}

\title{State-Aware Proximal Pessimistic Algorithms  for Offline Reinforcement Learning
}

\author[1]{Chen Chen \thanks{Corresponding author,
E-mail address: cclvr@163.com.}
}
\author[2]{Hongyao Tang}
\author[2]{Yi Ma}
\author[1]{Chao Wang}
\author[3]{Qianli Shen}
\author[4]{Dong Li}
\author[2]{Jianye Hao}
\affil[1]{Qiyuan Lab}
\affil[2]{College of Intelligence and Computing, Tianjin University}
\affil[3]{NUS, Singapore}
\affil[4]{Noah’s Ark Lab, Huawei}

\begin{document}
\maketitle

\begin{abstract}
Pessimism is of great importance in offline reinforcement learning (RL). One broad category of  offline RL algorithms fulfills pessimism by explicit or implicit behavior regularization.
However, most of them only consider policy divergence as behavior regularization, ignoring the effect of how the offline state distribution differs with that of the learning policy, which may lead to under-pessimism for some states and over-pessimism for others. Taking account of this problem, we propose a principled algorithmic framework for offline RL, called \emph{State-Aware Proximal Pessimism} (SA-PP). The key idea of SA-PP is leveraging discounted stationary state distribution ratios between the learning policy and the offline dataset to modulate the degree of behavior regularization in a state-wise manner, so that pessimism can be implemented in a more appropriate way. We first provide theoretical justifications on the superiority of SA-PP over
previous algorithms, demonstrating that SA-PP produces a lower suboptimality upper bound in a broad range of settings. Furthermore, we propose a new algorithm named \emph{State-Aware Conservative Q-Learning} (SA-CQL), by building SA-PP upon representative CQL algorithm with the help of DualDICE for estimating discounted stationary state distribution ratios. Extensive experiments on standard offline RL benchmark show that SA-CQL  outperforms the popular baselines on a large portion of benchmarks and attains the highest average return.
\end{abstract}


\section{Introduction}
\label{introduction}
Reinforcement learning (RL) has  achieved considerable success in many decision making and control domains, 
such as Game Playing \cite{Mnih2015DQN,SilverHMGSDSAPL16AlphaGO,vinyals2019grandmaster}, Robotics Manipulation \cite{HafnerLB020Dream,Lillicrap2015DDPG,Smith19AVID},
Medicine Discovery \cite{Popova19Molecule,schreck2019retrosyn,YouLYPL18GCPN} and so on.
However, the expensive  online interaction cost prevents RL from being applied  into practice, thus it is  crucial to make full use of the data collected previously in an offline manner, which is  the core topic in  offline RL ~\cite{levine2020offline}.

The key challenge of offline RL is the disastrous value overestimation especially for unfamiliar states and actions,  mainly caused by the distribution shift between the offline dataset and the state-action distribution induced by the learning policy.
It further leads the policy optimization towards  an unexpected or even destructive direction \cite{fujimoto2019off,levine2020offline}.
A major solution to this issue is the use of pessimism principle \cite{buckman2020importance,liu2020provably,jin2021pessimism, xie2021bellman,rashidinejad2021bridging,zanette2021provable,kumar2021should}, which resorts to pessimistic value estimates
to eliminate the negative impact of unreliable estimation.
One broad category  of practical offline RL methods fulfills pessimism by behavior regularization  \cite{kumar2019stabilizing,fujimoto2019off,wu2019behavior,kostrikov2021offline,kumar2020conservative,fujimoto2021minimalist}, which is typically done  by augmenting the critic or actor loss with a penalty measuring the divergence of the learning policy from the behavior policy.
While being sound and effective,
these algorithms only consider how the learning policy differs from the behavior policy (i.e., action-aware),
ignoring the influence of the discrepancy in state distribution between the learning policy and the offline dataset.
We consider that
action-aware pessimism alone is deficient, and taking account of the state occupancy in offline dataset as well as that of the learning policy
is important to offline RL.

\begin{figure}
\begin{center}
\centerline{\includegraphics[width=0.6\columnwidth]{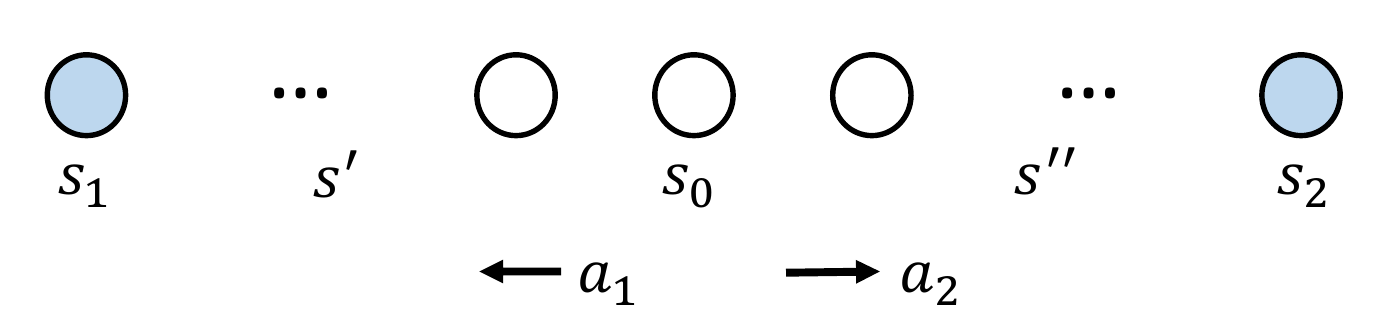}}
\vspace{0.2cm}
\caption{A motivating example for the significance of state-aware pessimism.}
\label{fig:intro}
\end{center}
\vspace{-0.8cm}
\end{figure}

We provide a motivating example in Figure \ref{fig:intro} to better illustrate the deficiency of the policy divergence-based regularization.
Consider a chain MDP with initial state $\bs_0$ and two absorbing states $\bs_1$ and $\bs_2$.
First assume that the offline dataset $\mathcal{D}$ contains $m$ trajectories from $\bs_0$ to $\bs_1$ and $n$ trajectories from $\bs_0$ to $\bs_2$ with $n \gg m$.
As a result,  for some $\bs'$ lying between $\bs_0$ and $\bs_1$ and $\bs''$ lying between $\bs_0$ and $\bs_2$,
$\bs'$ is less familiar to the agent than $\bs''$ due to its lower state occupancy.
According to the pessimism principle,
more pessimism is ought to be made when estimating the value of $\bs'$.
This indicates that the pessimism degree should be $\propto \frac{1}{d^{\mathcal{D}}(s)}$ where $d^{\mathcal{D}}(s)$ denotes the state density in $\mathcal{D}$.
However, for a uniform learning policy $\pi$ with $\pi(\ba_1|\bs)  = \pi(\ba_2|\bs) = 0.5$ for all $s$,  the policy divergence-based regularization typically imposes equal degree of pessimism to $\bs'$ and $\bs''$,
since the
empirical behavior policy $\hatbehavior$ on $\bs'$ and $\bs''$ is calculated  as $\hatbehavior(\ba_1|\bs') = 1, \hatbehavior(\ba_2|\bs') = 0$, and $\hatbehavior(\ba_1|\bs'') = 0, \hatbehavior(\ba_2|\bs'') = 1$.
Such improper pessimism can be implemented for many possible learning policies,
which inevitably results in  over-pessimistic value estimations on some states as well as under-pessimistic value estimations on some others.
Moreover, 
to evaluate the performance of the learning policy $\pi$, the states visited frequently by  $\pi$  deserve  more reliable value estimations. Assume that $\pi$ visits some $s_1$ very frequently, then the estimated return from some initial distribution $\rho$, $\hat{V}(\rho) = \sum_{s}d_{\pi}(s)\sum_a\pi(a|s)\hat{r}(s,a)$ is affected by $d_{\pi}(s)$,with $d_{\pi}(s)$ the state distribution induced by $\pi$,  and thus the uncertainty of $\hat{r}(s_1,a)$ may be further enlarged due to the high proportion  $d_{\pi}(s)$. To avoid the disastrous overestimation induced by the high occupancy of $s_1$,
it is reasonable  to be more pessimistic on the estimated reward of $s_1$, $\hat{r}(s,a)$,  which is naturally equivalent to more pessimistic value estimates of $s_1$. 
This can be characterized by letting pessimism degree be $\propto d_{\pi}(s)$ under the pessimism principle.

In an overall view,   policy divergence based regularization is insufficient to fulfill the pessimism principle well and we consider that a more appropriate \textit{state-aware} pessimism should be $\propto \frac{d_{\pi}(s)}{d^{\mathcal{D}}(s)}$. Note that \cite{liu2020provably} shares a similar motivation that the effect of state distributions needs to be taken account of when implementing pessimism, but it realizes the idea  by  constraining bellman backups on a support set $\{(s,a): d^{\mathcal{D}}(s,a) > b\}$ with $b$ a predefined hyper-parameter, thus  is in fact a binary pessimism modulation rather than a finely modulated pessimism as we expected.

To this end, we propose a principled algorithmic framework, called \emph{State-Aware Proximal Pessimism} (SA-PP).
The core idea of SA-PP is leveraging the stationary state distribution ratios between the learning policy and the offline dataset (i.e., $\frac{d_{\pi}(s)}{d^{\mathcal{D}}(s)}$), 
to finely modulate the pessimism degree of  behavior regularization in a state-wise manner.
In this way, insufficient or excessive pessimism induced by typical behavior regularization can be compensated by the ratios. Theoretically,  we conduct elaborate analysis for the composite impact of both overestimation and underestimation on the suboptimality, to compare the treatments  with and without the state-aware modulation, and prove that SA-PP is  prone to generate lower suboptimality upper bound.  We also provide more practical  conditions under which SA-PP is superior over its counterpart for both small and large conservative weights cases, demonstrating that the superiority of SA-PP holds for a board range of settings. Furthermore,  we extend the ratios $\frac{d_{\pi}(\bs)}{d^{\D}(\bs)}$ to $f(\frac{d_{\pi}(\bs)}{d^{\D}(\bs)})$ with $f$ a monotonically increasing real function so that it can be adopted in practice with better flexibility. 

In general,  SA-PP  can be regarded as a flexible plug-in component for many behavior regularization based offline RL algorithms.
For a representative instance,
we propose a practical algorithm called \textit{State-Aware Conservative Q-Learning} (SA-CQL),
by building SA-PP upon the behavior regularization adopted in Conservative Q-Learning \cite{kumar2020conservative}.
Specifically, 
we use DualDICE \cite{nachum2019dualdice} to estimate the ratios $\frac{d_{\pi}(s)}{d^{\mathcal{D}}(s)}$ , which are  used to re-weight the conservative value estimation of CQL, to realize a state-aware modulation of behavior regularization.
Extensive experiments on standard offline RL benchmark D4RL \cite{d4rl} and Atari datasets released in \cite{agarwal2020optimistic} show that SA-CQL achieves  the state-of-the-art  on most datasets, and  outperforms existing behavior   regularization  based methods with a remarkable improvement on some datasets.  This suggests that behavior regularization based offline RL can significantly benefit from the state-aware pessimism.


\vspace{-0.03cm}

\section{Background}
For common notations, we follow the convention.
For two vectors $\boldsymbol{x}, \boldsymbol{y}$, $\langle \boldsymbol{x}, \boldsymbol{y} \rangle$ means the inner product of $\boldsymbol{x}$ and $\boldsymbol{y}$, $\boldsymbol{x}/\boldsymbol{y}, \boldsymbol{x}\cdot \boldsymbol{y}$ and  $f(\boldsymbol{x})$ are all element-wise operations, $\boldsymbol{x}(i)$ means the $i$'th element of $\boldsymbol{x}$.
For a space $X$, we use $|X|$ to denote the dimensionality and $Dist(X)$ to denote all possible probability distributions over $X$.

\vspace{-0.03cm}
\paragraph{Reinforcement Learning}

Consider a Markov decision process (MDP) defined by a tuple $\mdp := (\states, \actions, \transitions, r, \gamma, \rho)$,
with the state space $\states$, the action space $\actions$, the transition function $\transitions: S \times A \to Dist(S)$, the reward function $r: S \times A \to \mathbb{R}$, the discount factor $\gamma \in [0,1)$ and the initial state distribution $\rho$.
Without loss of generality, we consider a bounded reward function $r(\bs,\ba) \in [-1,1]$ for convenience in our theoretical derivation.
An RL agent interacts with the MDP by its policy $\pi: \states \rightarrow Dist(\actions)$,
with the aim of maximizing the expected cumulative discounted reward (or expected discounted return), i.e., $\mathbb{E}_{\pi} \left[\sum_{t=0}^{\infty}\gamma^{t} r(\bs_t, \ba_t) \right]$ with $\bs_0 \sim \rho(\cdot)$, $\ba_{t} \sim \pi(\cdot | \bs_{t})$, $\bs_{t+1} \sim P \left( \cdot \mid \bs_{t}, \ba_{t} \right)$.

Denote $\joint$ as the joint space of $\states \times \actions$, 
we use $\boldsymbol{r} \in \real^{|\joint|}$ to represent the vector of reward function, and similarly
use $\transitions \in \real ^{|\joint| \times |\states|}$ for the dynamics matrix and $\boldsymbol{\rho} \in \real^{|\states|}$ for the vector of initial state distribution, with some reuse of notation.
We then follow \cite{buckman2020importance} to define an \textit{activity matrix} $A^{\pi} \in \real^{|\states|\times |\joint|}$ for each policy $\pi$, 
which encodes the state-conditional state-action distribution of $\pi$, by letting $A^{\pi}(\bs, \langle \tilde{\bs}, \ba \rangle) = \pi(\ba|\bs)$ if $\bs = \tilde{\bs} $ otherwise $A^{\pi}(\bs, \langle \tilde{\bs}, \ba \rangle) = 0$.
It is used to represent the dynamics of policy $\pi$ acting in the MDP by $A^{\pi} \transitions \in \real^{|\states|\times |\states|}$ or $\transitions A^{\pi} \in \real^{|\joint|\times |\joint|}$.
Thus, for any two state $\bs, \bs'$, the probability of being in $\bs'$ after $t$ time
steps when starting from $\bs$ and following policy $\pi$ is $[(A^{\pi}\transitions)^t](\bs, \bs')$.
Furthermore, it can be derived that $\sum_{t=0}^{\infty} (\gamma A^{\pi}P)^t = (I_{|\states|}-\gamma A^{\pi}P)^{-1}$. The marginal discounted state distribution under $\pi$ starting from $\boldsymbol{\rho}$ is denoted by $\freq_{\rho}^{\pi}(\bs) = \boldsymbol{\rho} (I_{|\states|}-\gamma A^{\pi}P)^{-1} (\bs)$,
simplified as $\freq^{\pi}(\bs)$ in the rest of this paper.

For any policy $\pi \in \Pi$, 
its state-action value function $Q^{\pi}: \states \times \actions \rightarrow \real$ is defined as the expected cumulative discounted reward obtained by $\pi$ for any $\bs,\ba$:
$Q^{\pi}(\bs,\ba) = \mathbb{E}_{\pi} \left[\sum_{t=0}^{\infty}\gamma^{t} r(\bs_t, \ba_t)|\bs_0=\bs, \ba_0=\ba \right]$.
Similarly, the state value function is defined as, $V^{\pi}(\bs) = \mathbb{E}_{\pi} \left[\sum_{t=0}^{\infty}\gamma^{t} r(\bs_t, \ba_t)|\bs_0=\bs \right]$.
Most off-policy RL algorithms follow the paradigm of (general) Policy Iteration \cite{SuttonB98RLAI},
which alternates (partial) policy evaluation and (approximate) policy improvement.
In policy evaluation,
the value function of $\pi$ is evaluated, typically
by iterating the Bellman operator as $\mathcal{B}^\pi Q= \boldsymbol{r} + \gamma \transitions A^{\pi} Q$ or $\mathcal{B}^\pi V= A^\pi \boldsymbol{r} + \gamma A^{\pi} \transitions V$,
with the fixed point $Q^{\pi}$ and $V^{\pi}$ respectively.
Note that $\transitions A^{\pi} Q(\bs, \ba) = \E_{\bs' \sim \transitions(\bs' | \bs, \ba), \ba' \sim \pi(\ba'|\bs')} \left[ Q(\bs', \ba') \right]$.
While in policy improvement, the policy
$\policy(\ba|\bs)$ is improved by updating towards actions that maximize the estimated $Q$-values. 

\vspace{-0.03cm}
\paragraph{Offline Reinforcement Learning}

In offline RL, online interaction is no longer allowed and instead, we consider an offline dataset $\mathcal{D}$, which is generated by some unknown state distribution $d^{\mu}$ and behavior policy $\behavior$(or by multiple ones).
For ease of expression,
we use $n_{\D}(\bs), n_{\D}(\bs,\ba), n_{\D}(\bs,\ba, \bs')$ to denote the numbers of state $\bs$, pair $(\bs,\ba)$ and transition $(\bs,\ba, \bs')$ in $\D$. 
We further define $n^{-1/2}_{\D, \pi}(\bs) := \E_{\ba \sim \pi( a\cdot|\bs)} \left[ n_{\D}(\bs,\ba)^{-1/2} \right]$ for later use.
We use $\boldsymbol{{n}}_{\D}$ and $\boldsymbol{{n}^{-1/2}_{\D,\pi}}$ for corresponding
vectors respectively. 

Usually, empirical models are calculated with the samples in $\mathcal{D}$:
for any
state $\bs \in \mathcal{D}$, let
$\hatbehavior(\ba|\bs) := \frac{n_{\D}(\bs,\ba)}{n_{\D}(\bs)}$, 
$d^{\mathcal{D}}(\bs) := \frac{n_{\D}(\bs)} {|\mathcal{D}|}$ 
denote the empirical behavior policy and the empirical state distribution, respectively.
We use 
$\transitions_{\D}(\bs' | \bs, \ba ) := \frac{n_{\D}(\bs,\ba, \bs')}{n_{\D}(\bs,\ba)}$, 
$r_{\D}(\bs, \ba) := \frac{\sum_{\tilde{\bs}, \tilde{\ba} \in \mathcal{D}} r(\bs,\ba) \boldsymbol{1}[\tilde{\bs} = \bs, \tilde{\ba} = \ba]}{n_{\D}(\bs,\ba)}$ 
for the empirical transition function and empirical reward function,
yielding an empirical MDP $\mdp_{\D} := (\states, \actions, \transitions_{\D}, r_{\D}, \gamma, \rho)$.
Further, the policy evaluation step in $\D$ then can be defined by an empirical Bellman operator with $\mdp_{\D}$.
We denote this operator $\hat{\bellman}^\policy$ with $\hat{\bellman}^\policy Q = \boldsymbol{r}_{\D} + \gamma P_{\D} A^{\pi} Q$.
In turn, the marginal discounted state distribution under $\pi$ starting from $\boldsymbol{\rho}$ in $\mdp_{\D}$ is then calculated as  $\freq_{\rho, \D}^{\pi}(\bs) = \boldsymbol{\rho} (I_{|\states|}-\gamma A^{\pi}P_{\D})^{-1} (\bs)$, 
simplified as $\freq^{\pi}_{\D}(\bs)$.
We remind the reader of the difference between the notions $\freq^{\mathcal{D}}$ and $\freq_{\mathcal{D}}^{\pi}$ for correct understanding of our work introduced below.

\vspace{-0.03cm}
\paragraph{Proximal Pessimistic Policy Optimization}
One major class of existing offline RL algorithms follows the principle of pessimism and adopts behavior regularization
in policy evaluation to mitigate destructive overestimation. 
A family of such algorithms, called \emph{Proximal Pessimistic Algorithms}, is recently defined in a general form \cite{buckman2020importance} as follows:
\begin{definition}
A \emph{proximal pessimistic policy evaluation } algorithm $\mathscr{E}_{\text{Dis}}$,  is any algorithm in the family defined by the fixed-point function:
\begin{align}
\mathscr{E}_{\text{Dis}}(\pi,{\mathcal{D}}):= \bigg\{v \bigg|v = A^{\pi}(r_{\mathcal{D}} + \gamma P_{\mathcal{D}}v ) -\alpha \big( \textbf{Dis}(\pi,\hatbehavior) \big) \bigg\}, \nonumber
\end{align}
where $\textbf{Dis}(\pi,\hatbehavior) \in \real^{|\states|}$ is a state-wise distance vector between $\pi$ and $\hatbehavior$.
A \emph{proximal pessimistic policy optimization algorithm}
with subroutine  $\mathscr{E}_{\text{Dis}}$ is any algorithm with the following structure:
\begin{align}
    \mathscr{O}_{\text{Dis}}({\mathcal{D}}) := \arg\max_{\pi}\E_{\rho}[\mathscr{E}_{\bf{\text{Dis}}}(\pi,{\mathcal{D}})]. \nonumber
\end{align}
\vspace{-0.3cm}
\end{definition}
A number of prior methods  instantiate
this approach with different choices of $\textbf{Dis}$, including KL \cite{fujimoto2019off,wu2019behavior}, MMD \cite{kumar2019stabilizing}, and other distances \cite{kostrikov2021offline,kumar2020conservative,fujimoto2021minimalist}.
To evaluate the performance of different algorithms, suboptimality is used as the measure:
\begin{definition}
Given an offline dataset $\mathcal{D}$, 
for any policy optimization algorithm $\mathscr{O}$, 
the \emph{suboptimality} is computed by taking the difference between the expected return of an
optimal policy and the learning policy under the initial state distribution $\rho$:
\begin{small}
\begin{align}
    \textbf{SUBOPT}(\mathscr{O}({\mathcal{D}})) = 
    \langle\rho,  V_{\mdp}^{\pi^{*}} - V_{\mdp}^{\pi^{\mathscr{O}(\D)}} \rangle,
\end{align}
\end{small}
where $\pi^{*}$ is the true optimum in $\mdp$ and $\pi^{\mathscr{O}(\D)}$ is the proximal optimum obtained by $\mathscr{O}({\mathcal{D}})$.
\end{definition}

\vspace{-0.03cm}

\section{State-Aware Proximal Pessimistic Framework}

In this section, we introduce the \emph{state-aware} pessimism into the \emph{proximal pessimistic} framework, inducing the \emph{state-aware proximal pessimistic} framework. To this end, define the ratio as $w^{\pi/\mathcal{D}}(\bs): = d_{\D}^{\pi}(\bs)/d^{\mathcal{D}}(\bs) $, and let $\boldsymbol{w^{\pi/\mathcal{D}}} \in \real^{|\states|}$ be the vector in $\states$.  
\begin{definition}\label{def:SADis}
A \emph{state-aware proximal pessimistic (SA-PP) policy evaluation }  algorithm coupled with $\textbf{Dis}(\pi, \hatbehavior)$ is any algorithm in the family defined by the fixed-point function:
\begin{small}
\begin{align}
 \mathscr{E}_{\text{SA-Dis}}(\pi,{\mathcal{D}}):=  \bigg\{v \bigg|v = A^{\pi}(r_{\mathcal{D}} + \gamma P_{\mathcal{D}}v ) 
 -\alpha \boldsymbol{w^{\pi/\mathcal{D}}} \cdot \big( \textbf{Dis}(\pi,\hatbehavior) \big) \bigg\}. \nonumber
\end{align}
\end{small}
A \emph{state-aware proximal pessimistic policy optimization} (SA-PP) with subroutine  $\mathscr{E}_{\text{SA-Dis}}$ is any algorithm with the following structure
\begin{small}
\begin{align}
    \mathscr{O}_{\text{SA-Dis}}({\mathcal{D}}) := \arg\max_{\pi}\E_{\rho}[\mathscr{E}_{\bf{\text{SA-Dis}}}(\pi,{\mathcal{D}})]. \nonumber
\end{align}
\end{small}
\end{definition}
The state-aware framework  additionally considers how the learning policy differs from the dataset from the aspect of state level,  while the original framework fails to address. Intuitively, larger ratios mean that the states are relatively less occupied or they are more likely to be visited by the current policy,  which is thus expected  to be more pessimistic to obtain   reliable value estimates. On the other hand, smaller ratios mean that the  states are occupied  relatively more or they are less visited by the current policy,  which is  expected  to be not that pessimistic. 
We will provide theoretical guarantee for its superiority in the following sections. 



\subsection{Theoretical Results}

The following theoretical results are derived in the tabular setting and all  the proofs can be found in Appendix. Our primary goal is to show  that 
SA-PP is prone to generate a lower suboptimality upper bound than its conterpart, that is, the following inequality holds:
\begin{small}
\begin{align}
   \textbf{GOAL:} ~~~ \textbf{SUBOPT-UB}(\mathscr{O}_{\text{SA-Dis}}({\mathcal{D}})) \leq \textbf{SUBOPT-UB}(\mathscr{O}_{\text{Dis}}({\mathcal{D}})). \label{proof_goal}
\end{align}
\end{small}
We firstly provide a lemma which is  obtained by making minor modifications to Theorem 4 in \cite{buckman2020importance} and upper bounding the uncertainty in the tabular setting. 
\begin{lemma}\label{lemma:FDPO}
For any
dataset $\mathcal{D}$ and a policy space $\Pi$, consider any proximal pessimistic  policy optimization algorithm $\mathscr{O}({\mathcal{D}})$ coupled with the proximal pessimistic term $\boldsymbol{p} \in \real^{|\states|}$ by  $\mathscr{O}_{\textbf{p}}({\mathcal{D}}) := \arg\max_{\pi}\E_{\rho}[\mathscr{E}_{\bf{\textbf{p}}}(\pi,{\mathcal{D}})]$  and $\alpha \in (0, 1)$ is a pessimism hyperparameter,  then  the suboptimality of $\mathscr{O}({\mathcal{D}})$ is bounded with probability at least $1-\delta$ by 
\begin{small}
\begin{equation}
\begin{aligned}
 \textbf{SUBOPT}(\mathscr{O}({\mathcal{D}})) 
& \leq \inf_{\pi \in \Pi} \big( \langle \rho, V^{\pi^{*}} - V^{\pi}\rangle 
+ \langle \boldsymbol{\freq}_{\D}^{\pi},C_0\boldsymbol{n}^{-1/2}_{\D,\pi} + \alpha \boldsymbol{p} \rangle  \big) + \sup_{\pi \in \Pi}    \big( \langle \boldsymbol{\freq}_{\D}^{\pi},C_0\boldsymbol{n}^{-1/2}_{\D,\pi} - \alpha \boldsymbol{p} \rangle\big), 
\label{lemma:upper_bound} 
\end{aligned}
\end{equation}
\end{small} where 
$C_0 = \frac{1}{1-\gamma}\min\bigg(\sqrt{\frac{1}{2}\ln(\frac{2|\mathcal{S}|\times |\mathcal{A}|}{\delta})}, \sqrt{\frac{1}{2}\ln(\frac{2|\mathcal{S}|\times |\Pi|}{\delta})}\bigg)$.

\end{lemma}
It is explained in \cite{buckman2020importance} that the two terms inside $\inf$ brackets capture the suboptimality and the underestimation errors for $\pi$ respectively, and the supremum term  corresponds to the largest overestimation error on any $\pi$. The upper bound in Lemma \ref{lemma:FDPO} is tight and reveals an asymmetry between the impact of overestimation errors and underestimation errors. This inspires us  that the overestimation error is much more crucial  to be avoided than the underestimation error, the lower overestimation error $\boldsymbol{p}$ introduces,  the lower overall error it may generates.  We will
formally prove this intuition in Theorem  \ref{thm:key_condition}.  To this end, denote the RHS of (\ref{lemma:upper_bound}) as $\textbf{SUBOPT-UB}(\mathscr{O}({\mathcal{D}})) $,  the terms inside $\inf\big(\cdot\big)$ and  $\sup \big( \cdot \big)$ in Equation (\ref{lemma:upper_bound}) with $\boldsymbol{p} = \textbf{Dis}(\pi,\hatbehavior)$ as $\textbf{INF}_{\textbf{Dis}}$, 
$\textbf{SUP}_{\textbf{Dis}}$, with $\boldsymbol{p} = \boldsymbol{w}^{\pi/\D}\cdot \textbf{Dis}(\pi,\hatbehavior)$ as 
$\textbf{INF}_{\textbf{SA-Dis}}$ and  $\textbf{SUP}_{\textbf{SA-Dis}}$ respectively. 
\vspace{0.3cm}
\begin{theorem} \label{thm:key_condition}
(\ref{proof_goal}) holds with probability $1-\delta$
given that
\begin{small}
\begin{align}
   &\langle \boldsymbol{d}^{\overline{\pi}_1}_{\D}, \big(\boldsymbol{d}^{\overline{\pi}_1}_{\D}/\boldsymbol{d}^{\D} - \boldsymbol{1}\big)\cdot \textbf{Dis}(\overline{\pi}_1, \hatbehavior)  \rangle 
   \langle \boldsymbol{d}^{\overline{\pi}_2}_{\D}, \big(\boldsymbol{d}^{\overline{\pi}_2}_{\D}/\boldsymbol{d}^{\D} - \boldsymbol{1}\big)\cdot \textbf{Dis}(\overline{\pi}_2, \hatbehavior)  \rangle, \label{thm:equ:key_condition}
\end{align}
\end{small}
where 

$
    \overline{\pi}_1 := \sup_{\pi \in \Pi} \textbf{SUP}_{\textbf{SA-Dis}}(\pi)$,  and $
\overline{\pi}_2  := \inf_{\pi \in \Pi}\textbf{INF}_{\textbf{Dis}}(\pi) 
$.

\end{theorem}
\vspace{0.3cm}
\begin{remark}
Intuitively, (\ref{thm:equ:key_condition}) is prone to be satisfied and some reasoning is put in Appendix.
\end{remark}

We now provide a more specific condition straightforward to be  verified such that  (\ref{thm:equ:key_condition}) is met with.  Some assumptions are required.
\begin{assumption}\label{ass:lower_bounds}
$\hatbehavior(\ba|\bs ) \geq \varepsilon_{\beta} > 0$, for $\bs, \ba \in \D$.
\end{assumption}
\begin{assumption}\label{ass:ergodic}
Define $\bs_1 = \arg\min_{\bs}\freq^{\D}(\bs)$, there exists a policy $\pi_0 \in \Pi$  such that  $\freq^{\pi_0}_{\D}(\bs_1) > \varepsilon_d \in (0, 1)$ and $\max_{\bs}\textbf{Dis}(\pi_0, \hatbehavior) \leq \Delta_{\beta}$. 
\end{assumption}

\begin{assumption} \label{ass:optimal_policy}
$\overline{\pi}_2$ satisfies $\freq_{\D}^{\overline{\pi}_2}(\bs)/\freq^{\D}(\bs) \leq 1+ c,, \forall \bs \in \states$, where $\overline{\pi}_2$ is defined in Theorem \ref{thm:key_condition} and $c >0$.
\end{assumption}

\begin{remark}\label{remark:assumption}
Assumption \ref{ass:lower_bounds} requires that $\hatbehavior$  puts enough probabilities on all supported actions. Assumption \ref{ass:ergodic} can be satisfied in many cases. For example, when the dynamics  $P_{\D}$ are deterministic, and there must exists
 a path from some $\bs$ to $\bs_1$ in $\D$, then  $\pi_0$  can be constructed such that the probabilities of these paths  are all $1$ under $\pi_0$, then $\varepsilon_d $ is positive at this time. From the way to construct $\pi_0$,  it can be deduced
straightforwardly that $\varepsilon_d $ is  intrinsic of the MDP problem and also independent of $\D$, so is $\Delta_{\beta}$ since $\pi_0$ is supported by $\hatbehavior$ according to $\pi_0$'s construction.
Assumption \ref{ass:optimal_policy} is indeed satisfied,
since $\overline{\pi}_2$ is the infimum of $\langle \boldsymbol{d}^{\pi}_{\D},C_0\boldsymbol{n}^{-1/2}_{\D,\pi} + \alpha \textbf{Dis}(\pi,\pi_{\beta})   \rangle$, then $\overline{\pi}_2$ should be covered by $\pi^{\beta}$ otherwise $\langle \boldsymbol{d}^{\pi}_{\D}, C_0 \boldsymbol{n}^{-1/2}_{\D,\pi} + \alpha \textbf{Dis}(\pi,\pi_{\beta})   \rangle$ will be positive infinite.
\end{remark}

\begin{theorem}\label{thm:small_alpha_plus}
Under Assumptions \ref{ass:lower_bounds}-\ref{ass:optimal_policy}, if $\alpha = \alpha'/|\D|$ satisfying  $\alpha' < C_0\varepsilon_d /\Delta_{\beta}   $ and the following conditions hold:
\begin{align}
  C'_{\mathcal{M}} (\frac{\varepsilon_{\beta}}{d^{\D}(\bs_1)} -\sqrt{\varepsilon_{\beta}}) \textbf{Dis}(\overline{\pi}_1, \hatbehavior)(\bs_1)  > (1+c), \label{thm:core_thm:condition}
\end{align}
where $ C'_{\mathcal{M}}$ is a constant independent of $\D$, then (\ref{proof_goal}) holds with probability $1-\delta$.
\end{theorem}


\begin{remark}
We can
always pick some appropriate $\alpha'$ satisfying $\alpha' < C_0\varepsilon_d /\Delta_{\beta}$ , since $C_0$, $\varepsilon_d$ and $\Delta_{\beta}$ are all intrinsic to the MDP problem and is also independent  of $\D$, as discussed  in Remark \ref{remark:assumption}. 
\end{remark}
\begin{remark}
 (\ref{thm:core_thm:condition})  demonstrates how the related factors are coupled together  and implies some  insight in the limiting cases. To further understand this point, 
let us consider a common setting that $\hat{\pi}_{\beta}$ and $d^{\D}$ are weakly coupled, where
  $\hat{\pi}_{\beta}$ is fixed merely leaving $d^{\D}$ as a variable of $\D$ (This can be achieved when $d^{\D}$ is not generated  by $\hat{\pi}_{\beta}$). If we further constrain the policy class to contain all the polices supported by $\hatbehavior$  with a lower bound $\varepsilon_{\beta}$, then $c$ can be upper bounded and $\textbf{Dis}(\overline{\pi}_1, \hatbehavior)(\bs_1)$ can be lower bounded as well by some simple derivations. At this time,  the LHS of  (\ref{thm:core_thm:condition})  tends infinity as $d^{\D}(\bs_1)$ tends to $0$, meaning that (\ref{thm:core_thm:condition}) holds for small enough $d^{\D}(\bs_1)$.  This exactly reveals a nice property that SA-PP is advantageous especially for the dataset with an extremely non-uniform state distribution. 
\end{remark}

Theorem \ref{thm:key_condition} and  \ref{thm:small_alpha_plus} both focus on the situation where $\alpha$ is not sufficient to rule out the uncertainty which yields a positive  supremum term in Equation (\ref{lemma:upper_bound}). In the case that $\alpha$ is large enough and there only remains the infimum term in (\ref{lemma:upper_bound}), though  Theorem \ref{thm:key_condition}-\ref{thm:small_alpha_plus} do not hold any more, SA-PP is still advantageous given that the ratios are clipped above by some value. Specifically, 
\begin{theorem}\label{thm:large_alpha}
Assume that  $\alpha > C_0 \max_{\bs}\big(\boldsymbol{w}^{\pi/\D}  \cdot \textbf{Dis-CQL}(\pi, \hatbehavior) \cdot \boldsymbol{n}^{-1/2}_{\D} \big)(\bs)$, then  there exists some value $C >1$ such that,  once that  $w^{\pi/\D}(\bs)$ is clipped above by $C$, (\ref{proof_goal}) holds  with probability $1-\delta$.
\end{theorem}

\begin{remark}
Theorem \ref{thm:large_alpha} is straightforward  since at this time, conservative weight $\alpha$ is sufficiently large to cancel out overestimation errors and the underestimation errors are left  as the main concern, then clipped ratios can  make the value estimation  not that pessimistic as before, which can effectively reduce the underestimation errors. 
\end{remark}

\subsection{Extension to $f$-State-Aware Proximal Pessimistic Algorithms}
Besides the original $w^{\pi/\D}(\bs)$, one may extend the above derivations to a more general class of state-aware pessimism with $f(w^{\pi/\D})$,  where $f: \real_{+} \rightarrow \real_{+}$ is a  monotonically increasing real function. We define the proximal pessimistic approaches tuned by $f(w^{\pi/\D})$ as \emph{ $f$-state-aware proximal pessimistic} ($f$-SA-PP) algorithms. 
   and Theorems \ref{thm:key_condition}, \ref{thm:small_alpha_plus} and \ref{thm:large_alpha}  for $f$-SA-PP can be extended to the corresponding forms respectively, see Appendix. 

It is  implied that the relative relationship  rather than the absolute values of the ratios are the key to achieve state-aware pessimism and a high-precision ratio estimation  is not demanded actually.
Such property is of great use in practice for controlling the range of  $w^{\pi/\D}(\bs)$ and making the iteration process more stable.

\subsection{State-Aware Conservative Q learning}
Based on the general SA-PP algorithm presented in previous section, we further derive a practical implementation called \emph{State-Aware Conservative Q-Learning } (SA-CQL) to instantiate it. 
To be specific, we adopt CQL distance  \cite{kumar2020conservative} mentioned previously $\textbf{Dis-CQL}(\pi, \hatbehavior)(\bs) = \E_{\pi(\cdot|\bs)}[\pi(\cdot|\bs)/\hatbehavior(\cdot|\bs) -1]$ as the behavior regularization, then the policy evaluation process becomes:
\begin{small}
\begin{align}
    \hat{V}^{k+1} = A^{\pi}(\boldsymbol{r}_{\mathcal{D}} + \gamma P_{\mathcal{D}}\hat{V}^{k})- \alpha \boldsymbol{w}^{\pi/\D} \cdot \textbf{Dis-CQL}(\pi,\hatbehavior), \forall k,
    \label{SACQL_update_V}
\end{align}
\end{small}
 which is  equivalent to, 
\begin{small}
\begin{align}
    \hat{Q}^{k+1}(\bs, \ba) = \hat{\bellman}^\policy \hat{Q}^{k}(\bs,\ba) - \alpha \frac{\freq_{\D}^{\pi}(\bs)}{\freq^{\D}(\bs)} \frac{\pi(\ba|\bs)-\hatbehavior(\ba|\bs)}{\hatbehavior(\ba|\bs)}, \forall \bs, \ba, k. \label{SACQL_update}
\end{align}
\end{small}

\textbf{Connection to CQL} Above Q iteration process (\ref{SACQL_update}) corresponds to the SA-CQL objective
\begin{small}
\begin{equation}
\begin{aligned}
    \hat{Q}^{k+1} \leftarrow &\arg\min_{Q}~~ \alpha  \bigg(\E_{\bs \sim \textcolor{red}{d_{\D}^{\pi}(\bs)}, \ba \sim \pi(\ba|\bs)}\left[Q(\bs, \ba)\right] 
   - \E_{\bs \sim \textcolor{red}{d_{\D}^{\pi}(\bs)}, \ba \sim \hatbehavior(\ba|\bs)}\left[Q(\bs, \ba)\right] \bigg) 
    + \text{TD-error},\label{sacql_objective}
\end{aligned}
\end{equation}
\end{small}rather than the original CQL objective 
\begin{small}
\begin{equation}
\begin{aligned}
    \hat{Q}^{k+1} &\leftarrow \arg\min_{Q}~~ \alpha \bigg(\E_{\bs \sim \textcolor{red}{d^{\mathcal{D}}(\bs)}, \ba \sim \pi(\ba|\bs)}\left[Q(\bs, \ba)\right] 
   - \E_{\bs \sim \textcolor{red}{d^{\mathcal{D}}(\bs)}, \ba \sim \hatbehavior(\ba|\bs)}\left[Q(\bs, \ba)\right] \bigg)
    + \text{TD-error}.\label{cql_objective}
\end{aligned}
\end{equation}
\end{small}It can be seen that SA-CQL differs from CQL on the state distribution  the expectation is based on, which implies that SA-CQL assigns conservativeness particularly on the states with respect to the learning policy instead of all the states in the dataset and the original form may induce excessive pessimism on highly occupied  or irrelevant states. 
The underestimation property for CQL is still maintained as follows:
\begin{theorem}[Underestimation results)]
For any $\pi(\ba|\bs)$ , with probability $1-\delta$, the value of the policy under the Q function from Equation (\ref{sacql_objective}), $\hat{V}^{\pi}(\bs) = \E_{\pi(a|s)}[\hat{Q}^{\pi}(\bs,\ba)]$ lower bounds the true value of the policy obtained via exact policy evaluation ${V}^{\pi}(s) = \E_{\pi(\ba|\bs)}[Q^{\pi}(\bs,\ba)]$, according to
\begin{small}
\begin{align}
\hat{V}^{\pi}(\bs) &\leq V^{\pi}(\bs) -\alpha w^{\pi/\D}(\bs) \textbf{Dis-CQL}(\pi, \hatbehavior)(\bs) + C_0\boldsymbol{n}^{-1/2}_{\D, \pi} (\bs). \nonumber
\end{align}
\end{small}
Thus, if 
$\alpha > C_0 \max_{\bs}\big(\boldsymbol{w}^{\pi/\D}  \cdot \textbf{Dis-CQL}(\pi, \hatbehavior) \cdot \boldsymbol{n}^{-1/2}_{\D} \big)(\bs) $,
we have $\hat{V}^{\pi}(s) \leq V^{\pi}(s),  \forall s \in \mathcal{D}^{\pi}$.
\end{theorem}

Practically, we adopt CQL($\mathcal{H}$) as the backbone and generate the optimization objective $\text{SA-CQL}(Q)$ to solve $Q$  at iteration $k$, which is:
\begin{small}
\begin{align}
  & \min_{Q}~ \alpha \E_{\bs \sim d^\D(\bs)}\omega^{\pi_k/\D}(\bs)\bigg[\log \sum_{\ba} \exp(Q(\bs, \ba)) 
     - \E_{\ba \sim \hatbehavior(\ba|\bs)}   \big[Q(\bs, \ba)\big]\bigg]+\frac{1}{2}\E_{\bs, \ba, \bs' \sim \mathcal{D}}\bigg[\big(Q - \hat{\bellman}^{\policy_k} \hat{Q}^{k} \big)^2 \bigg] , \label{eqn:practical_objective}
\end{align}
\end{small}
where $\omega^{\pi_k/\D}(\bs) = \frac{\freq^{\pi_k}(\bs,\ba)}{\freq^{\D}(\bs,\ba)}\frac{\hatbehavior(\ba|\bs)}{\pi^k(\ba|\bs)}$  and $\frac{\freq^{\pi_k}(\bs,\ba)}{\freq^{\D}(\bs,\ba)} =\zeta^{\pi_k/\D}(\bs, \ba)$ is estimated by 
solving the following  min-max saddle-point optimization problem \cite{nachum2019dualdice}.\begin{small}
\begin{equation}
\begin{aligned}
   & \min_{\nu}\max_{\zeta} J(\nu, \zeta)  :=  \E_{\bs,\ba,\bs'\in \D, \ba'\sim \pi_k(\bs')} \big[ (\nu(\bs,\ba) 
   -\gamma\nu(\bs',\ba')) \zeta(\bs,\ba)\nonumber \\ &-\zeta(\bs,\ba)^2/2 \big] 
  -(1-\gamma)\E_{\bs_0\sim \rho, \ba_0\sim \pi^k(\bs_0)} \big[ \nu(\bs_0,\ba_0) \big]. \label{equ:dualDICE}
\end{aligned}
\end{equation}
\end{small}It is analyzed in \cite{nachum2019dualdice} that the solution of Equation (\ref{equ:dualDICE}) exactly gives an estimate of the density ratio. 
See  Appendix.for the pseudo-code of SA-CQL and the discussion about the computation cost.


\vspace{-0.03cm}
\begin{figure}[!htbp]
		\begin{subfigure}{.5\textwidth}
			\centering
			\includegraphics[width=\textwidth]{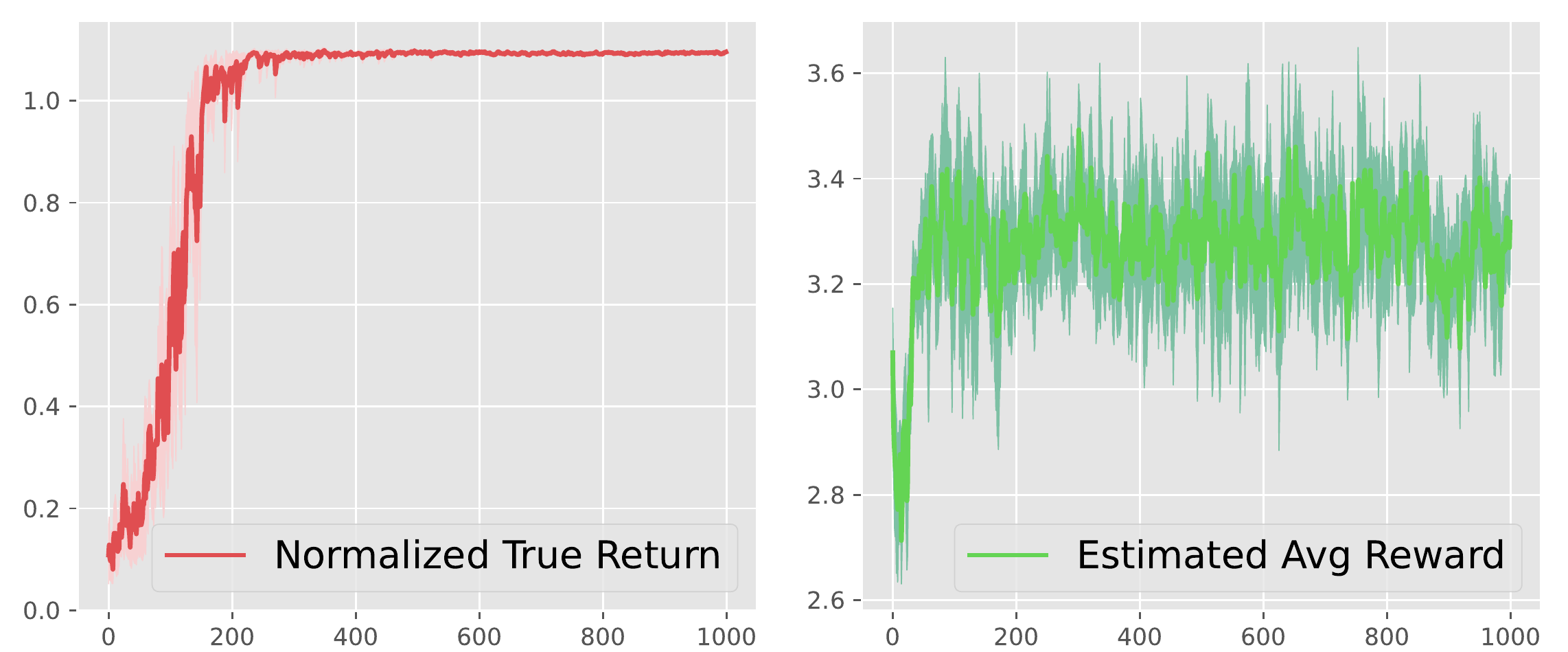}
			\caption{expert}
		\end{subfigure}
		\begin{subfigure}{.5\textwidth}
			\centering
			\includegraphics[width=\textwidth]{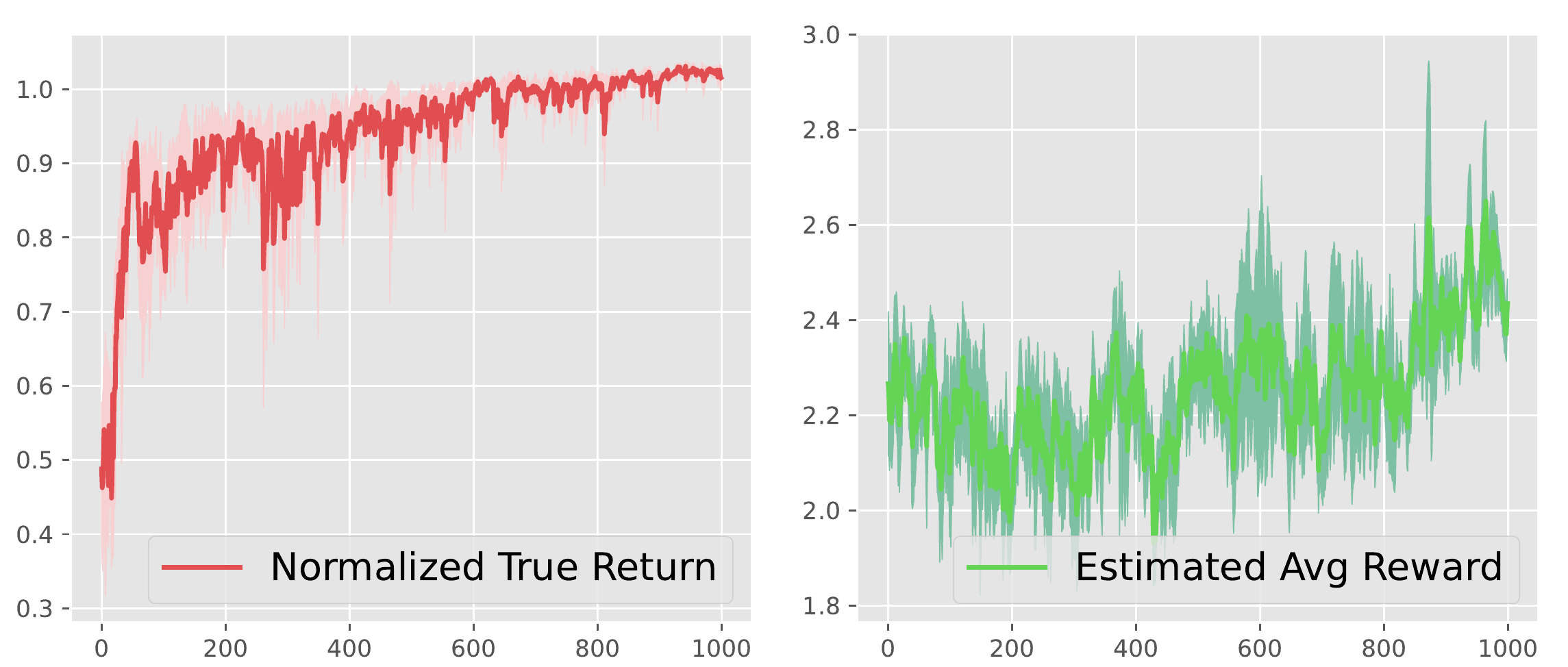}
			\caption{full-replay}
		\end{subfigure}
\caption{\footnotesize Evaluation of Stationary State Distribution Ratios}\label{fig:dice}
\end{figure}

\vspace{-0.03cm}

\section{Experimental Evaluation}
\label{sec:experiments}


\vspace{-0.05cm}
\subsection{Setups}
\label{subsec:setups}
We compare our proposed SA-CQL to prior offline RL methods on continuous control datasets from the D4RL benchmark~\cite{d4rl}, including three environments (halfcheetah, hopper, and walker2d) and six dataset types (medium, medium-replay, full-replay, expert, medium-expert, random).
We compare to prior offline RL algorithms that use different forms of behavior regularization, including:
 KL divergence based  BCQ~\cite{fujimoto2019off}, MMD based BEAR~\cite{kumar2019stabilizing},
 Fisher Divergence based F-BRC~\cite{kostrikov2021offline}, Euclidean distance-based  TD3PlusBC~\cite{fujimoto2021minimalist},
 value regularization-based 
CQL~\cite{kumar2020conservative}, 
as well as  two
uncertainty based algorithm, UWAC ~\cite{wu2021uncertainty}, EDAC ~\cite{an2021uncertainty}.
These baselines contain various choices of behavior regularization and  representative uncertainty-based algorithms, which fully reflect the state-of-the-art.   

Besides, we also evaluate discrete SA-CQL on offline, image-based Atari games \cite{bellemare2013arcade}. We compare SA-CQL to REM \cite{agarwal2020optimistic} and CQL \cite{kumar2020conservative} on the four Atari tasks (Pong, Breakout, Qbert, Seaquest), using the dataset released by the authors of \cite{agarwal2020optimistic}.

Specifically, we choose $f$-state ratio with $f(x) = b_1\cdot (\max_x \log x - \min_x \log x(\log x - \min_x \log x)+b_0$ with $b_0, b_1$ as the hyperparameters.
Implementation details are provided in Appendix.

\begin{table*}[!htbp]
\caption{\footnotesize Comparison of our method (SA-CQL) to prior work,  where `m',`m-r',`f-r',`e',`m-e' and `r' represent `medium', `medium-replay', `full-replay',`expert',`medium-expert' and `random', respectively.}

\scalebox{0.90}{
\begin{tabular}{lcclccccr}
\hline
\textbf{Dataset} & UWAC            & BEAR              & BCQ                 & EDAC                     & TD3PlusBC                & CQL                 & F-BRC              & SA-CQL                                                   \\ \hline
halfcheetah-m    & 42.0$\pm$ 0.47  & 42.0$\pm$ 0.1     & 47.7$\pm$ 0.6       & \bf{64.2$\pm$ 2.1}       & 48.8$\pm$ 0.3            & 52.5$\pm$ 0.3       & 48.3$\pm$ 0.5      & \bf{58.1$\pm$ 0.8}                                       \\
halfcheetah-m-r  & 36.1$\pm$ 4.4   & 36.7$\pm$ 1.8     & 44.7$\pm$ 0.9       & \bf{63.3$\pm$ 1.7}       & 44.6$\pm$ 0.7            & 49.3$\pm$ 0.3       & 43.9$\pm$ 1.9      & \bf{55.1$\pm$ 1.4}                                       \\
halfcheetah-f-r  & 62.3$\pm$ 2.3   & 62.5$\pm$ 1.9     & 74.5$\pm$ 1.5       & \bf{82.5$\pm$ 2.3}       & 74.3$\pm$ 2.6            & 80.5$\pm$ 0.5       & 71.4$\pm$ 3.2      & \bf{83.1$\pm$ 1.0}                                       \\
halfcheetah-e    & 92.5$\pm$ 0.7   & 91.6$\pm$ 0.4     & 96.6$\pm$ 1.9       & 4.8$\pm$1.1              & 97.1$\pm$ 0.4            & \bf{100.5$\pm$ 2.7} & 95.4$\pm$ 0.9      & \bf{98.2 $\pm$ 1.9}                                      \\
halfcheetah-m-e  & 42.95$\pm$ 0.3  & 45.3$\pm$4.0      & \bf{92.8$\pm$ 1.0}  & 72.2$\pm$32.6            & 83.4$\pm$2.4             & 68.5$\pm$ 12.       & \bf{94$\pm$ 0.8}   & 81.2$\pm$ 1.1                                            \\
halfcheetah-r    & 2.3$\pm$ 0.005  & 2.3$\pm$ 0.002    & 2.2$\pm$ 0.002      & \bf{28.4$\pm$ 0.3}       & 10.7$\pm$ 1.3            & 26.2$\pm$ 0.6       & 26.9$\pm$ 1.4      & \bf{31$\pm$ 0.8}                                         \\ \hline
hopper-m         & 49.7$\pm$ 7.4   & 49.5$\pm$ 3.1     & 63.3$\pm$7.9        & \bf{101.3$\pm$ 0.8}      & 60.5$\pm$ 3.4            & 74.1$\pm$ 4.6       & \bf{90.9$\pm$ 6.5} & 86.3$\pm$ 3.8                                            \\
hopper-m-r       & 30.8$\pm$ 13.1  & 37.2$\pm$ 0.4     & 25.4$\pm$ 1.3       & \bf{101.5$\pm$ 0.6}      & 53.4$\pm$ 17.8           & 98.3$\pm$ 1.7       & 93.1$\pm$ 1.6      & \bf{100.1$\pm$ 3.3}                                      \\
hopper-f-r       & 21.9$\pm$ 7.9   & 69.5$\pm$ 15.8    & 34.6$\pm$ 4.5       & 106.1$\pm$ 0.1           & 89$\pm$ 13.8             & \bf{107.3$\pm$ 0.3} & 106$\pm$ 1.9       & \bf{108$\pm$ 0.8}                                        \\
hopper-e         & 111.4$\pm$ 0.8  & 78.4$\pm$ 6.5     & 59.9$\pm$5.9        & 29.8$\pm$16.7            & 108.4$\pm$ 3.6           & \bf{112.1$\pm$} 0.8 & 111.5$\pm$ 0.4     & \bf{111.8$\pm$ 0.6}                                      \\
hopper-m-e       & 50.9$\pm$ 7.8   & 49.7$\pm$ 1.4     & 112.4$\pm$ 0.8      & 88.1$\pm$ 32.3           & 102.0$\pm$ 6.5           & \bf{103$\pm$ 7}     & 101.3$\pm$ 8.7     & \bf{111.8$\pm$ 0.9}                                      \\
hopper-r         & 2.6$\pm$ 0.1    & 7.2$\pm$ 0.3      & 7.3$\pm$ 0.3        & 7.7$\pm$ 0.3             & 8.9$\pm$ 0.3             & \bf{12.1$\pm$ 2.9}  & 11.03$\pm$ 17.1    & \bf{17.7$\pm$ 12.5}                                      \\ \hline
walker2d-m       & 78.3$\pm$ 2.8   & 74.6$\pm$8.6      & 77.3$\pm$ 3.4       & \bf{89.8$\pm$0.4}        & 85.0$\pm$0.4             & 85.4$\pm$ 0.7       & 81.7$\pm$ 1.7      & \bf{87.7$\pm$ 0.5}                                       \\
walker2d-m-r     & 25.5$\pm$ 7.1   & 5.4$\pm$5.4       & 34.3$\pm$ 17.2      & 81.7$\pm$0.1             & 84.2$\pm$5               & 82.9$\pm$ 2.6       & \bf{86.6$\pm$ 2.5} & \bf{90.1$\pm$ 3.1}                                       \\
walker2d-f-r     & 25.6$\pm$ 31.2  & 93.3$\pm$2.6      & 57.4$\pm$ 12.7      & \bf{98.6$\pm$1.2}        & 94.6$\pm$1.4             & 97.7$\pm$ 1.7       & 96.7$\pm$ 0.8      & \bf{102.3$\pm$ 1.3}                                      \\
walker2d-e       & 108.0$\pm$ 0.6  & 105.8$\pm$6.0     & \bf{110.7$\pm$ 1.3} & 37.7$\pm$51.1            & 110$\pm$0.2              & \bf{110.4$\pm$ 0.6} & 108.9$\pm$ 0.3     & \bf{109.2$\pm$ 0.3}                                      \\
walker2d-m-e     & 107.16$\pm$ 2.8 & 108.3$\pm$ 2.1    & 109.6$\pm$ 0.9      & \bf{113.9$\pm$0.4}       & 110$\pm$ 0.4             & 109.6$\pm$ 0.4      & 109.5$\pm$ 0.2     & \bf{109.8$\pm$ 0.3}                                      \\
walker2d-r       & 2.8$\pm$ 0.2    & \bf{4.9$\pm$ 0.5} & 4.3$\pm$ 1.5        & 0.$\pm$ 0.0              & 1.8$\pm$ 0.9             & 0.                  & 2.4$\pm$ 3.8       & \bf{4.1$\pm$ 0.2}                                        \\ \hline
Average          & 49.6            & 53.6              & 58.61               & \multicolumn{1}{c}{65.1} & \multicolumn{1}{c}{70.4} & 75.5                & 76.6               & \multicolumn{1}{c}{{\color[HTML]{F56B00} \textbf{80.3}}} \\ \hline
\end{tabular}\label{total-resault-table}
}
\end{table*}


\subsection{Comparison with Baselines}
The results of our method and all
considered baselines on D4RL benchmark are presented  in Table \ref{total-resault-table} and we highlight the top-2 best results in bold on each dataset. It can be observed that our method achieves the top-2 best on almost all  datasets, and attains  the highest average score among all methods. In particular, SA-CQL outperforms CQL, which is being on top of,  on almost all the non-expert datasets by a remarkable margin, suggesting that behavior regularization based offline RL methods can significantly benefit from state-aware pessimism. It  deserves to be noted that although EDAC  also reaches the top-2 best on several datasets as SA-CQL does, its computation cost is significantly larger than SA-CQL due to the usage of a large number of ensembles and detailed comparison is put in the Appendix. 

The results on Atari tasks are shown in Table \ref{total-resault-atari-table}, which also demonstrates the performance gain of SA-CQL against the discrete baselines, especially on Qbert.

\begin{table}[htbp]
\centering
\caption{Comparison of our method (SA-CQL) to prior work on Atari tasks}\label{total-resault-atari-table} 
\begin{tabular}{|l|l|l|l|}
\hline
         & REM            & CQL             & SA-CQL          \\ \hline
Pong     & 10.1 $\pm$ 3.3 & 15.65 $\pm$ 2.2 & \bf{16.9 $\pm$ 2.2}  \\ \hline
Qbert    & 6778 $\pm$ 248 & 13844 $\pm$ 736 & \bf{17562 $\pm$ 972} \\ \hline
Seaquest & 1523 $\pm$ 345 & 1147 $\pm$ 408  & \bf{1667 $\pm$ 677}  \\ \hline
Breakout & 81 $\pm$ 2.5   & 93 $\pm$ 4.2    & \bf{99 $\pm$ 3.1}    \\ \hline
\end{tabular}
\end{table}

\vspace{-0.03cm}

\subsection{Ablation Study}
\textbf{Evaluation of   State Distribution Ratios.} It is challenging to verify the quality of the state distribution ratios  due to the lack of ground truth, and thus we turn to compare  the true return with the estimated average reward using the ratios as done in \cite{nachum2019dualdice}, to see whether the ratios are reliable or not. We conduct experiments on two datasets of walker2d, and  Figure \ref{fig:dice} shows that the estimated average reward demonstrates a similar upward trend 
as the true return, which means that the learning ratios  are informative indicators of true density ratios.
We also remark that in fact the high-precision estimations are not strictly required  since  $f$-SA-PP framework implies that the relative relationships instead of the absolute values are sufficient to obtain a satisfactory performance.

\begin{figure}[!htbp]
\centering
        \begin{minipage}{0.3\textwidth}
            \centering
            \includegraphics[width=1\textwidth]{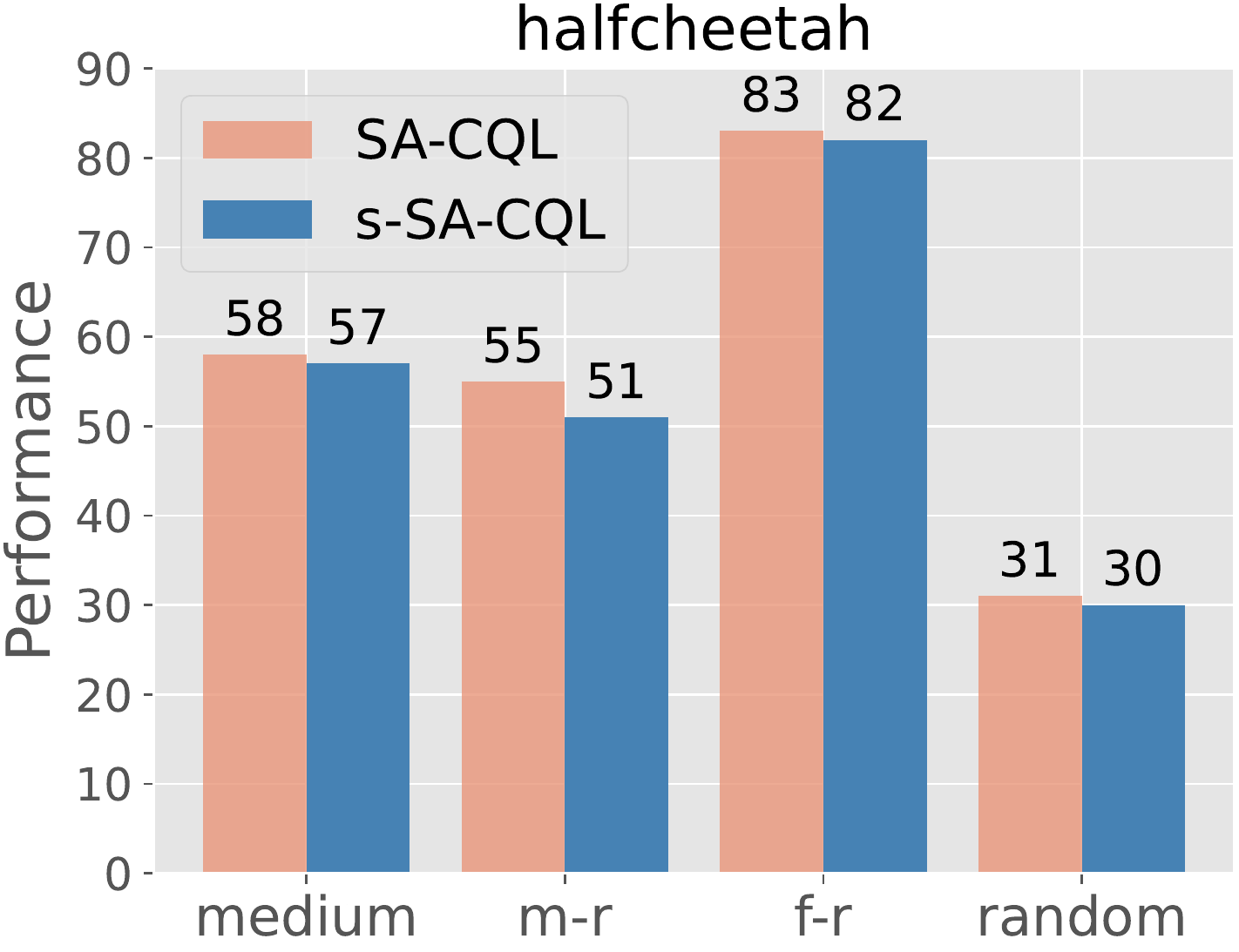}
        \end{minipage}
        \begin{minipage}{0.3\textwidth}
            \centering
            \includegraphics[width=1\textwidth]{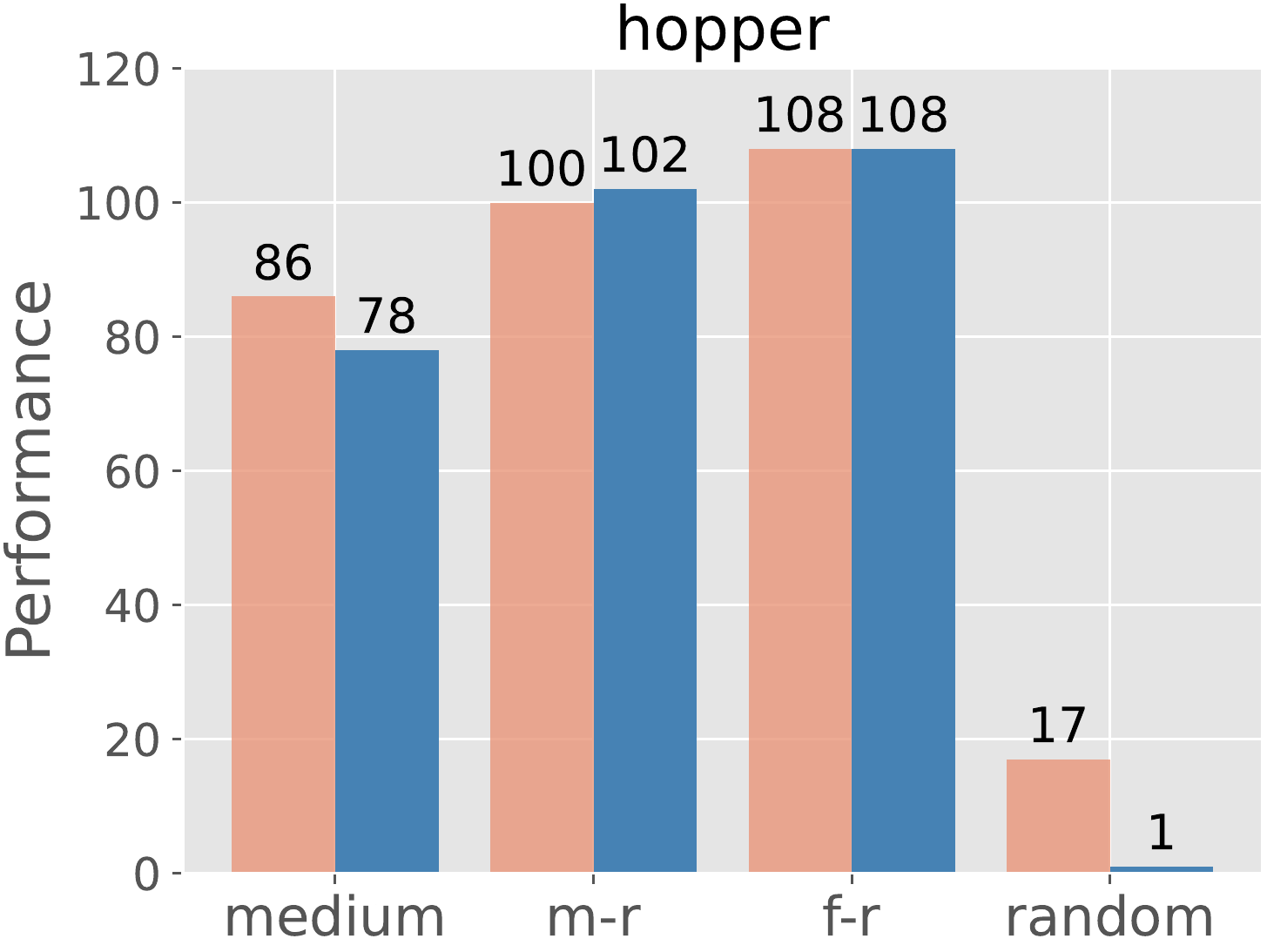}
        \end{minipage}
        \begin{minipage}{0.3\textwidth}
            \centering
            \includegraphics[width=1\textwidth]{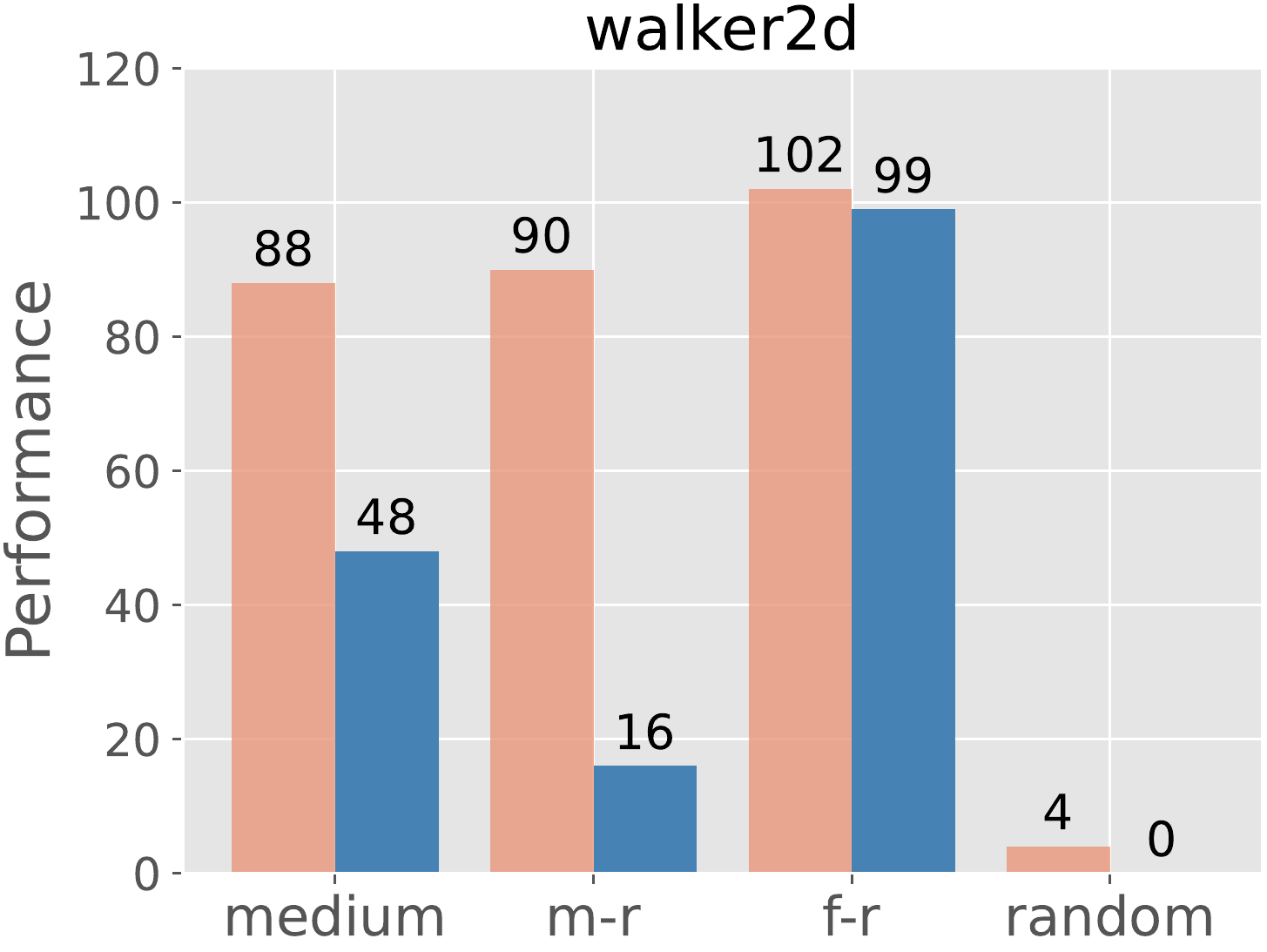}
        \end{minipage}
  \caption{\footnotesize Comparison of SA-CQL and simplified SA-CQL. }
  \label{fig:ablation_ratio} 
    \end{figure}

\vspace{-0.01cm}
\textbf{Impact of  State Distribution Ratio.} In order to guarantee  that the state distribution ratios do work, we conduct an ablation study to compare SA-CQL with  a simplified version of SA-CQL (s-SA-CQL) with the  ratio replaced by  a random value between $(b_0, b_1)$.  It can be observed in Figure \ref{fig:ablation_ratio} that  s-SA-CQL can achieve close or even slightly better performance with SA-CQL on some datasets, which verifies Theorem \ref{thm:large_alpha} in some degree that clipped state-aware pessimism can reduce excessive pessimism whenever conservative weight is large enough. But on other datasets, especially on the hardest task walker2d, s-SA-CQL fails to  complete the task with extremely bad performance, showing that the  state distribution ratios do reflect the appropriate state-aware pessimism and play an irreplaceable role in improving the performance.

\vspace{-0.03cm}
\textbf{SA-CQL vs CQL with different Conservativeness.} Since we set the upper bound of  the  state distribution ratios $b_1$ as $5$ for some datasets, which may make SA-CQL more conservative than CQL due to the composite effect of $b_1 \cdot \alpha$. 
To guarantee that CQL cannot be improved only by changing conservative weight, or say, SA-CQL outperforms  due to state-aware pessimism rather than tricky hyper-parameter setting, we conduct another ablation study  to compare SA-CQL with CQL using different $\alpha$. The results  demonstrated in Appendix. Table 6 show that
SA-CQL  still outperforms the best CQL baseline on almost all datasets and remarkably outperforms it on half of the datasets. This reveals  state-aware pessimism is the necessity for performance improvement.

\vspace{-0.03cm}
\section{Related Work}\label{related_work}
Offline RL algorithms are
especially prone to fail due to erroneous value estimation
induced by the distributional shift between the dataset and the learning policy. Pessimism is key to the success of offline RL algorithms.  In the theoretical line, it is proved that pessimistic value iteration can alleviate overestimation effectively and achieve good performance even with non-perfect data coverage,  which is typically done by adding uncertainty-based pessimism \cite{buckman2020importance,jin2021pessimism,liu2020provably,kumar2021should,rashidinejad2021bridging}, represented by the quantifier which can upper bound the errors of empirical Bellman operators,  or proximal pessimism \cite{buckman2020importance}, represented by the policy divergence, as a penalty term into the policy evaluation process.   Besides,  ``global pessimism" \cite{xie2021bellman,zanette2021provable} is introduced which only implements pessimism in the initial state rather than in all states in a point-wise way. We are  inspired by the pessimistic value iteration framework,
but focus more on the comparison between proximal pessimism and state-wise proximal pessimism, by conducting intensive analysis for the composite impact of both overestimation and underestimation on the suboptimality.  What is more, SA-PP is also practical to implement with extensive empirical evaluations, which remarkably differs from existing theoretical works. 

In the algorithmic line, there  are broadly two categories of offline RL methods: uncertainty based ones and behavior regularization based ones,  which can be viewed respectively as the instantiation of uncertainty-based pessimism and proximal pessimism mentioned above to some extent. Uncertainty based approaches attempt to estimate the epistemic uncertainty of
Q-values or dynamics, and then utilize this uncertainty to pessimistically estimating Q in a model-free manner \cite{agarwal2020optimistic,wu2021uncertainty,an2021uncertainty}, or  conduct learning on the pessimistic dynamic model  in a model-based manner \cite{yu2020mopo, kidambi2020morel}. This class of methods generally require multiple ensembles  to estimate the uncertainty and may induce a huge burden of computation and memory cost. Behavior regularization based algorithms constrain the learned policy to lie close to the behavior policy   in either explicit or implicit ways. \cite{kumar2019stabilizing,fujimoto2019off,wu2019behavior,kostrikov2021offline,kumar2020conservative,fujimoto2021minimalist}, and is advantageous over uncertainty based methods in computation efficiency and memory consumption.    The implementations primarily vary in  the  choice of behavior regularizer: KL \cite{fujimoto2019off,wu2019behavior} , MMD \cite{kumar2019stabilizing}, and others \cite{kumar2020conservative, kostrikov2021offline,fujimoto2021minimalist}. SA-PP adds to this class of approaches yet further considers the effect of stationary state distribution ratios and thus overcomes the limitation of typical behavior regularization that pessimism cannot be implemented appropriately. Moreover, our method
can be integrated on top of existing methods straightforwardly and thus maintains their practical advantages. 

There  still exists a family of offline RL methods which leverages the regularized policy optimization framework rather than the typical value based framework to learn the optimal policy \cite{nachum2019algaedice,lee2021optidice, xu2021offline,lee2020representation, shen2021model}.  The divergence of  stationary state-action distributions, instead of policies,  are imposed as behavior regularization, which plays a similar role as our state-aware pessimism. But SA-PP is built upon the classical dynamic programming based RL framework with rich theoretical guarantee and experimental success, and  can be implemented directly on top of any pessimistic value-based approaches. Nevertheless, we still resort to the superiority of regularized policy optimization framework in the field of off-policy estimation  \cite{nachum2019dualdice}, and make high quality estimates of ratios so that SA-PP is feasible in practice.

\section{Conclusion and Future Work}\label{sec:conclusion}
We have proposed the SA-PP framework, which is built upon the behavior regularization-based offline RL approaches, and  utilizes the discounted stationary state distribution ratios between the learning policy and the dataset to state-wisely modulate the degree of the behavior regularization. Theoretical justifications on the superiority of SA-PP over the methods without state-aware modulation are provided, showing that SA-PP produces lower suboptimality upper bound compared with its conterparts. SA-PP is also instantiated as SA-CQL on top of CQL, with the discounted stationary state distribution ratios estimated by the DualDICE algorithm. Empirical experiments show that SA-CQL achieves superior performance on offline RL benchmarks with remarkable improvements over existing popular methods, demonstrating the importance of state-aware pessimism.

There still exists some future work to do: current theoretical results are established in the tabular setting, which is worthy of being extended to the continuous setting. The concept of the state-aware pessimism should also be combined with other behavior regularization-based offline RL methods and evaluated in more complex tasks to further validate its effectiveness.






%
\bibliography{references}





\onecolumn
\appendix

\section{Proofs}\label{app:proof}
\subsection{Proof of Lemma \ref{lemma:FDPO}}\
In order to prove Lemma \ref{lemma:FDPO}, we provide some uncertainty function  first. Follow Definition 4 in \cite{buckman2020importance}, \begin{definition}\label{appendix:lemma:uncertainty}
A function $\boldsymbol{u}_{\D,\delta}: \states \times \actions \rightarrow \R$ is a state-action-wise Bellman uncertainty function, if for a dataset $\D$ it obeys with probability at least $1-\delta$ for all $\pi$ and $Q$.
\begin{align}
    \boldsymbol{u}_{\D,\delta} \geq |\hat{\mathcal{B}^{\pi}}Q  -\mathcal{B}^{\pi}Q |. \nonumber
\end{align}
A function $\boldsymbol{u}^{\pi}_{D,\delta}: \states  \rightarrow \R$ is a state-wise Bellman uncertainty function, if for a dataset $\D$ it obeys with probability at least $1-\delta$ for all $\pi$ and $ Q$.
\begin{align}
\boldsymbol{u}_{\D,\delta}^{\pi}(\bs) \geq |A^{\pi}(\hat{\mathcal{B}^{\pi}}Q -\mathcal{B}^{\pi}Q )| . \nonumber
\end{align}
A function $\boldsymbol{\mu}^{\pi}_{D,\delta}: \states  \rightarrow \R$ is a value Bellman uncertainty function, if for a dataset $\D$ it obeys with probability at least $1-\delta$ for all $\pi$ and $ Q$.
\begin{align}
\boldsymbol{\mu}_{\D,\delta}^{\pi} \geq \langle \boldsymbol{d}^{\pi}_{\D},  |A^{\pi}(\hat{\mathcal{B}^{\pi}}Q  -\mathcal{B}^{\pi}Q )| \rangle. \nonumber
\end{align}
\end{definition}

\begin{lemma}[state-action-wise bound, B.1 in \cite{buckman2020importance}]\label{state-action-bound}
In tabular setting, we have
\begin{align}
 |A^{\pi}(\hat{\mathcal{B}^{\pi}}Q  -\mathcal{B}^{\pi}Q )| \leq \frac{1}{1-\gamma} \min \bigg( \bigg( \sqrt{\frac{1}{2}\ln \frac{2 |\states \times \actions|}{\delta}} \bigg) \boldsymbol{n}^{-1/2}_{\D,\pi},1  \bigg) \nonumber
\end{align}
\end{lemma}

\begin{lemma}[state-wise bound, B.2 in \cite{buckman2020importance}]\label{state-bound}
\begin{align}
 |A^{\pi}(\hat{\mathcal{B}^{\pi}}Q  -\mathcal{B}^{\pi}Q )| \leq \frac{1}{1-\gamma} \min \bigg( \bigg( \sqrt{\frac{1}{2}\ln \frac{2 |\states \times \Pi|}{\delta}} \bigg) \boldsymbol{n}^{-1/2}_{\D,\pi},1  \bigg) \nonumber
\end{align} 
\end{lemma}
The above bounds have different forms due to the different ways to apply Hoeffding’s inequality. The first bound is obtained by invoking Hoeffding’s inequality at each of the $|\states \times \actions|$ state-actions and taking a union bound.  The second bound is obtained by invoking Hoeffding’s inequality at each of the $|\states|$ states and $\Pi$ policies, and taking a union bound. Please refer to  \cite{buckman2020importance} for some details.

\textbf{Proof of Lemma\ref{lemma:FDPO}}
By Theorem 4 in \cite{buckman2020importance} and replace the total variation there by $\boldsymbol{p}$, we have 
\begin{small}
\begin{align}
\textbf{SUBOPT}(\mathscr{O}({\mathcal{D}})) 
\leq \inf_{\pi \in \Pi}\bigg ( \langle \rho, V^{\pi^{*}} - V^{\pi}\rangle 
&+ \boldsymbol{\mu}^{\pi}_{\D,\delta} + \langle \boldsymbol{\freq}_{\D}^{\pi} ,\alpha \boldsymbol{p} \rangle  \bigg) + \sup_{\pi \in \Pi}    \big( \boldsymbol{\mu}^{\pi}_{\D,\delta}- \langle \boldsymbol{\freq}_{\D}^{\pi}, \alpha \boldsymbol{p} \rangle\big), \label{appendix:lemma:p}
\end{align}
\end{small}
where $\boldsymbol{\mu}^{\pi}_{\D,\delta}$ is the \emph{value uncertainty function} defined in Definition \ref{appendix:lemma:uncertainty}.  
By Lemma \ref{state-action-bound}-\ref{state-bound}, we can pick some $\boldsymbol{u}^{\pi}_{\D,\delta} $ such that
\begin{align}
    \boldsymbol{u}^{\pi}_{\D,\delta}  \leq C_0 \boldsymbol{n}^{-1/2}_{\D,\pi}, ~~~~
    C_0 := \frac{1}{1-\gamma}\min \bigg( \bigg( \sqrt{\frac{1}{2}\ln \frac{2 |\states \times \actions|}{\delta}} \bigg), \bigg( \sqrt{\frac{1}{2}\ln \frac{2 |\states \times \Pi|}{\delta}} \bigg)   \bigg). \nonumber
\end{align}
Then $\boldsymbol{\mu}^{\pi}_{\D,\delta} \leq \langle \boldsymbol{\freq}^{\pi}_{\D}, \boldsymbol{u}^{\pi}_{\D,\delta} \rangle \leq \langle \boldsymbol{\freq}^{\pi}_{\D}, C_0\boldsymbol{n}^{-1/2}_{\D,\pi} \rangle$.
By some simple arithmetic in (\ref{appendix:lemma:p})
  we can see Lemma \ref{lemma:FDPO} holds. \qed

\vspace{0.5cm}
\subsection{Proof of Theorem \ref{thm:key_condition}}
\textbf{Proof of Theorem \ref{thm:key_condition}}
By the definition of $\overline{\pi}_1$,  $\overline{\pi}_1$ takes the maximum of $\textbf{SUP}_{\textbf{SA-Dis}}(\pi)$, then we have 
\begin{align}
    \sup_{\pi \in \Pi} \textbf{SUP}_{\textbf{Dis}}(\pi) - \sup_{\pi \in \Pi} \textbf{SUP}_{\textbf{SA-Dis}}(\pi) 
    & \geq \textbf{SUP}_{\textbf{Dis}}(\overline{\pi}_1) - \textbf{SUP}_{\textbf{SA-Dis}}(\overline{\pi}_1) \nonumber \\
    & =  \alpha \cdot\langle \boldsymbol{d}^{\overline{\pi}_1}_{\D}, \big(\boldsymbol{d}^{\overline{\pi}_1}_{\D}/\boldsymbol{d}^{\D} - \boldsymbol{1}\big)\cdot \textbf{Dis}(\overline{\pi}_1, \hatbehavior)  \rangle . \label{proof: equ1}
\end{align}
By the definition of $\overline{\pi}_2$, $\overline{\pi}_2$ takes the minimum of $\textbf{INF}_{\textbf{Dis}}(\pi)$, we have 
\begin{align}
    \inf_{\pi \in \Pi} \textbf{INF}_{\textbf{SA-Dis}}(\pi) - \inf_{\pi \in \Pi} \textbf{INF}_{\textbf{Dis}}(\pi) 
    & \leq \textbf{INF}_{\textbf{SA-Dis}}(\overline{\pi}_2) - \textbf{INF}_{\textbf{Dis}}(\overline{\pi}_2) \nonumber \\
    & =  \alpha \cdot\langle \boldsymbol{d}^{\overline{\pi}_2}_{\D}, \big(\boldsymbol{d}^{\overline{\pi}_2}_{\D}/\boldsymbol{d}^{\D} - \boldsymbol{1}\big)\cdot \textbf{Dis}(\overline{\pi}_2, \hatbehavior)  \rangle . \label{proof: equ2}
\end{align}

Then combine (\ref{proof: equ1}) and (\ref{proof: equ2}) and by the condition (\ref{thm:equ:key_condition}), we have
\begin{align}
   & ~ \sup_{\pi \in \Pi} \textbf{SUP}_{\textbf{SA-Dis}}(\pi) + \inf_{\pi \in \Pi} \textbf{INF}_{\textbf{SA-Dis}}(\pi) -\big( \sup_{\pi \in \Pi} \textbf{SUP}_{\textbf{Dis}}(\pi)  +  \inf_{\pi \in \Pi} \textbf{INF}_{\textbf{Dis}}(\pi)  \big) \nonumber \\
   & \leq \alpha \bigg( \langle \boldsymbol{d}^{\overline{\pi}_2}_{\D}, \big(\boldsymbol{d}^{\overline{\pi}_2}_{\D}/\boldsymbol{d}^{\D} - \boldsymbol{1}\big)\cdot \textbf{Dis}(\overline{\pi}_2, \hatbehavior)  \rangle - \langle \boldsymbol{d}^{\overline{\pi}_1}_{\D}, \big(\boldsymbol{d}^{\overline{\pi}_1}_{\D}/\boldsymbol{d}^{\D} - \boldsymbol{1}\big)\cdot \textbf{Dis}(\overline{\pi}_1, \hatbehavior)  \rangle \bigg) \leq 0,
\end{align}
which yields  Theorem \ref{thm:key_condition}. \qed
 
\subsection{Discussion on Theorem \ref{thm:key_condition}} \label{discussion}
 (\ref{thm:equ:key_condition}) is prone to be satisfied due to the following analysis: since  $\overline{\pi}_2$ takes an infimum over $\Pi$, it will maintain a small suboptimality error as well as a small  $g(\overline{\pi}_2) :=\langle \boldsymbol{\freq}_{\D}^{\overline{\pi}_2},C_0\boldsymbol{n}^{-1/2}_{\D,\overline{\pi}_2} + \alpha  \textbf{Dis}\rangle$. Considering the fact  that $\boldsymbol{n}^{-1/2}_{\D,\pi}$ is proportional to $\boldsymbol{n}^{-1/2}_{\D}$ for the $\pi$ supported by $\hat{\pi}^{\beta}$, meaning that the first term of $g(\overline{\pi}_2)$ cannot be large, then   $ \boldsymbol{\freq}_{\D}^{\overline{\pi}_2}/\boldsymbol{d}^{\D}$ is not possible to be large as  well , otherwise the overall  $g(\overline{\pi}_2)$ cannot be small, this further yields a relatively small RHS of (\ref{thm:equ:key_condition}) . On the contrary
, $\overline{\pi}_1$ takes a supremum over $\Pi$,  and its induced state distribution tends to put most mass on the states with sparse occupancy in $\D$, then it inclines to generate a relatively large $ \boldsymbol{\freq}_{\D}^{\overline{\pi}_1}/\boldsymbol{d}^{\D}$ and also a large  LHS of  (\ref{thm:equ:key_condition}). 

In an overall view,   SA-PP does produce a smaller overestimation error,  yet at the cost of  increment of the underestimation error, but the reduction  of the overestimation error
(the LHS of (\ref{thm:equ:key_condition})) are prone to  counteract the increment of underestimation error (the RHS of (\ref{thm:equ:key_condition})). As a consequence, Theorem \ref{thm:key_condition}  shows that SA-PP is superior to its countepart for  a broad range of  settings.

\subsection{Proof of Theorem \ref{thm:small_alpha_plus}}
\textbf{Proof of Theorem \ref{thm:small_alpha_plus}}
By Assumption \ref{ass:ergodic}, we can find some $\pi_0$ with $\freq^{\pi_0}_{\D}(\bs_1) > \varepsilon_d$. 
Since $\overline{\pi}_1$ takes the maximum of $\textbf{SUP}_{\textbf{SA-Dis}}(\pi)$, we obtain
\begin{align}
    \langle \boldsymbol{\freq}_{\D}^{\overline{\pi}_1},C_0\textbf{n}^{-1/2}_{\D,\overline{\pi}_1}\rangle
    - \alpha'\frac{1}{|\D|}\langle \boldsymbol{\freq}_{\D}^{\overline{\pi}_1}, \frac{\boldsymbol{\freq}_{\D}^{\overline{\pi}_1}}{\boldsymbol{d}^{\D}}\textbf{Dis}(\overline{\pi}_1, \hatbehavior)\rangle 
     \geq \langle \boldsymbol{\freq}_{\D}^{\pi_0},C_0\textbf{n}^{-1/2}_{\D,\pi_0}\rangle
    - \alpha'\frac{1}{|\D|}\langle \boldsymbol{\freq}_{\D}^{\pi_0}, \frac{\boldsymbol{\freq}_{\D}^{\pi_0}}{\boldsymbol{d}^{\D}}\textbf{Dis}(\pi_0, \hatbehavior)\rangle, \nonumber 
\end{align}
which means
\begin{align}
 \langle \boldsymbol{\freq}_{\D}^{\overline{\pi}_1},C_0\textbf{n}^{-1/2}_{\D,\overline{\pi}_1}\rangle &\geq  \langle \underbrace{\boldsymbol{\freq}_{\D}^{\pi_0},C_0\textbf{n}^{-1/2}_{\D,\pi_0}\rangle}_{\uppercase\expandafter{\romannumeral1}} + \underbrace{\alpha'\frac{1}{|\D|}\langle \boldsymbol{\freq}_{\D}^{\overline{\pi}_1}, \frac{\boldsymbol{\freq}_{\D}^{\overline{\pi}_1}}{\boldsymbol{d}^{\D}}\textbf{Dis}(\overline{\pi}_1, \hatbehavior)\rangle - \alpha'\frac{1}{|\D|}\langle \boldsymbol{\freq}_{\D}^{\pi_0}, \frac{\boldsymbol{\freq}_{\D}^{\pi_0}}{\boldsymbol{d}^{\D}}\textbf{Dis}(\pi_0, \hatbehavior)\rangle}_{\uppercase\expandafter{\romannumeral2}}
 \label{app:equ:4}
\end{align}
For the first term, we have
\begin{align}
\uppercase\expandafter{\romannumeral1} \geq C_0 &\cdot\freq_{\D}^{\pi_0}(\bs_1)\sum_{\ba\in \hat\behavior(\cdot|\bs)}\frac{\pi_0(\ba|\bs)}{\sqrt{n_{\D}(\bs_1,a)}}  \nonumber \\
&\geq C_0\cdot\freq_{\D}^{\pi_0}(\bs_1)\sum_{\ba\in \hat\behavior(\cdot|\bs_1)}\frac{\pi_0(\ba|\bs_1)}{\sqrt{n_{\D}(\bs_1)}}  \geq C_0 \cdot\freq_{\D}^{\pi_0}(\bs_1)/\sqrt{n_{\D}(\bs_1)}.
\end{align}
The first inequality adopts Assumption \ref{ass:ergodic} that $\pi_0(\cdot|\bs_1)$ put all mass on the actions in $\D$.

For the second term, we have
\begin{align}
\uppercase\expandafter{\romannumeral2} =  \alpha'\langle
\big(\boldsymbol{\freq}_{\D}^{\overline{\pi}_1} \big)^2 \textbf{Dis}(\overline{\pi}_1, \hatbehavior) - \big(\boldsymbol{\freq}_{\D}^{\pi_0} \big)^2 \textbf{Dis}(\pi_0, \hatbehavior), \boldsymbol{n}^{-1}_{\D} \rangle
\geq - \alpha' \Delta_{\beta}/n_{\D}(\bs_1),
\end{align}
where the inequality holds due to Assumption \ref{ass:ergodic} that the policy divergence of $\pi_0$ is upper bounded by $\Delta_{\beta}$.

Combining $\uppercase\expandafter{\romannumeral1}$ and $\uppercase\expandafter{\romannumeral2}$ and ( \ref{app:equ:4}) we can obtain
\begin{align}
  \langle \boldsymbol{\freq}_{\D}^{\overline{\pi}_1},C_0\textbf{n}^{-1/2}_{\D,\overline{\pi}_1}\rangle & \geq    C_0 \cdot\freq_{\D}^{\pi_0}(\bs_1)/\sqrt{n_{\D}(\bs_1)} -\alpha' \Delta_{\beta}/n_{\D}(\bs_1)  \nonumber \\
 &\geq  C_0 \cdot\varepsilon_d/\sqrt{n_{\D}(\bs_1)} -\alpha' \Delta_{\beta}/n_{\D}(\bs_1). \label{app:equ:plus4}
\end{align}

Besides, by using Assumption \ref{ass:lower_bounds}, we have for $\forall \bs$,
\begin{align}
    n^{-1/2}_{\D,\overline{\pi}_1}(\bs) = n^{-1/2}_{\D}(\bs) \sum_{\ba} \frac{\overline{\pi}_1(\ba|\bs)}{\sqrt{\hatbehavior(\ba|\bs)}}  < n^{-1/2}_{\D}(\bs) \frac{1}{\sqrt{\varepsilon_{\beta}}}, 
\end{align}
which induces
\begin{align}
     \langle \boldsymbol{\freq}_{\D}^{\overline{\pi}_1},C_0\textbf{n}^{-1/2}_{\D,\overline{\pi}_1}\rangle \leq \langle \boldsymbol{\freq}_{\D}^{\overline{\pi}_1},C_0\textbf{n}^{-1/2}_{\D}\rangle \cdot \frac{1}{\sqrt{\varepsilon_{\beta}}}. \label{app:equ:5}
\end{align}

Combine (\ref{app:equ:plus4}) and (\ref{app:equ:5}) we obtain
\begin{align}
    \langle \boldsymbol{\freq}_{\D}^{\overline{\pi}_1},C_0\textbf{n}^{-1/2}_{\D}\rangle \geq \sqrt{\varepsilon_{\beta}}\frac{C_0 \cdot \varepsilon_d - \alpha'\Delta_{\beta}}{\sqrt{n_{\D}(\bs_1)}}, 
\end{align}
which reveals that 
$\freq_{\D}^{\overline{\pi}_1}(\bs_1) \geq (C_0 \cdot \varepsilon_d - \alpha'\Delta_{\beta}) \sqrt{\varepsilon_{\beta}} := C_{\mathcal{M}}\sqrt{\varepsilon_{\beta}}$, otherwise $\langle \freq_{\D}^{\overline{\pi}_1},C_0\textbf{n}^{-1/2}_{\D}\rangle$ would be decreased.

Then the LHS of (\ref{thm:equ:key_condition}) is 
\begin{align}
    &~~\langle \boldsymbol{d}^{\overline{\pi}_1}_{\D}, \big(\boldsymbol{d}^{\overline{\pi}_1}_{\D}/\boldsymbol{d}^{\D} - \boldsymbol{1}\big)\cdot \textbf{Dis}(\overline{\pi}_1, \hatbehavior)  \rangle \nonumber \\
    & \geq d^{\overline{\pi}_1}_{\D}(\bs_1)(\frac{d^{\overline{\pi}_1}_{\D}(\bs_1)}{d^{\D}(\bs_1)} -1) \textbf{Dis}(\overline{\pi}_1, \hatbehavior)(\bs_1) + \sum_{\bs \in \states/\bs_1} d^{\overline{\pi}_1}_{\D}(\bs)(0-1)\text{Dis}(\overline{\pi}_1, \hatbehavior)(\bs)\nonumber \\
    & \geq C_{\mathcal{M}}\sqrt{\varepsilon_{\beta}} (\frac{C_{\mathcal{M}}\sqrt{\varepsilon_{\beta}}}{d^{\D}(\bs_1)} -1) \textbf{Dis}(\overline{\pi}_1, \hatbehavior)(\bs_1) - (1-C_{\mathcal{M}}\sqrt{\varepsilon_{\beta}} ) \Delta_{\beta} \nonumber \\
    & \geq C_{\mathcal{M}}\sqrt{\varepsilon_{\beta}} (\frac{C_{\mathcal{M}}\sqrt{\varepsilon_{\beta}}}{d^{\D}(\bs_1)} -1) \textbf{Dis}(\overline{\pi}_1, \hatbehavior)(\bs_1) - \Delta_{\beta} \nonumber 
    \label{app:euq:final}
\end{align}
Without loss of generality, we can assume that $C_{\mathcal{M}} > 1$, then if the following inequality holds,
\begin{align}
  C'_{\mathcal{M}} (\frac{\varepsilon_{\beta}}{d^{\D}(\bs_1)} -\sqrt{\varepsilon_{\beta}}) \textbf{Dis}(\overline{\pi}_1, \hatbehavior)(\bs_1)  > (1+c),
\end{align}
 where $C'_{\mathcal{M}} = \frac{ C_{\mathcal{M}}^2}{\Delta_{\beta}}$, we will have
\begin{align}
   &~\langle \boldsymbol{d}^{\overline{\pi}_2}_{\D}, \big(\boldsymbol{d}^{\overline{\pi}_2}_{\D}/\boldsymbol{d}^{\D} - \boldsymbol{1}\big)\cdot \textbf{Dis}(\overline{\pi}_2, \hatbehavior)  \rangle 
     \leq c\Delta_{\beta} \nonumber \\
    & \leq \langle \boldsymbol{d}^{\overline{\pi}_1}_{\D}, \big(\boldsymbol{d}^{\overline{\pi}_1}_{\D}/\boldsymbol{d}^{\D} - \boldsymbol{1}\big)\cdot \textbf{Dis}(\overline{\pi}_1, \hatbehavior)  \rangle,
\end{align}
which yields Theorem \ref{thm:key_condition}. \qed



\subsection{Proof of Theorem \ref{thm:large_alpha}}
\textbf{Proof of Theorem \ref{thm:large_alpha}}
It is straightforward that $\textbf{INF}_{\textbf{SA-Dis}}(\pi) < \textbf{INF}_{\textbf{Dis}}(\pi)$ if $w^{\pi/\D} (\bs)$ is clipped above by $C=1$. Then for the special  $\overline{\pi} := \inf_{\pi \in \Pi}\textbf{INF}_{\textbf{Dis}}(\pi)$, we can increase $C$ slowly until  a critical $C > 1$ such that $\textbf{INF}_{\textbf{SA-Dis}}(\overline{\pi}) = \textbf{INF}_{\textbf{Dis}}(\overline{\pi})$ if $w^{\pi/\D} (\bs)$ is clipped above by $C$. Then $\inf_{\pi \in \Pi}\textbf{INF}_{\textbf{SA-Dis}}(\pi) \leq \textbf{INF}_{\textbf{Dis}}(\overline{\pi}) = \inf_{\pi \in \Pi}\textbf{INF}_{\textbf{Dis}}(\pi)$ when $w^{\pi/\D} (\bs)$ is clipped above by $C$.  \qed

\section{Extension to f -State-Aware Proximal Pessimistic Algorithms}\label{app:f_extenstion}
We define the proximal pessimistic approaches tuned by $f(w^{\pi/\D})$ as \emph{ $f$-state-aware proximal pessimistic} ($f$-SA-PP) algorithms. The goal is to prove that 
 \begin{align}
    \textbf{SUBOPT-UB}(\mathscr{O}_{\text{f-SA-Dis}}({\mathcal{D}})) < \textbf{SUBOPT-UB}(\mathscr{O}_{\text{Dis}}({\mathcal{D}})).  \label{proof_goal-f}
\end{align}
Theorems \ref{thm:key_condition}, \ref{thm:small_alpha_plus} and  \ref{thm:large_alpha}  for SA-PP can be extended as follows:

\begin{theorem} \label{thm:f-key_condition}
(\ref{proof_goal-f}) holds
if and only if
\begin{align}
   \langle \boldsymbol{d}^{\overline{\pi}_1}_{\D}, \big(f\big(\boldsymbol{d}^{\overline{\pi}_1}_{\D}/\boldsymbol{d}^{\D} \big)- \boldsymbol{1}\big)\cdot \textbf{Dis}(\overline{\pi}_1, \hatbehavior)  \rangle  
   \geq \langle \boldsymbol{d}^{\overline{\pi}_2}_{\D}, \big(f\big(\boldsymbol{d}^{\overline{\pi}_2}_{\D}/\boldsymbol{d}^{\D} \big)- \boldsymbol{1}\big)\cdot \textbf{Dis}(\overline{\pi}_2, \hatbehavior)  \rangle, \label{thm:f-equ:key_condition}
\end{align}
where $ \overline{\pi}_1 := \sup_{\pi \in \Pi} \textbf{SUP}_{\textbf{f-SA-Dis}}(\pi), ~~~
\overline{\pi}_2  := \inf_{\pi \in \Pi}\textbf{INF}_{\textbf{Dis}}(\pi)$.
\end{theorem}

\begin{assumption} \label{ass:optimal_policy-f}
$\overline{\pi}_2$ satisfies $f(\freq_{\D}^{\overline{\pi}_2}(\bs)/\freq^{\D}(\bs)) \leq 1+ c,, \forall \bs \in \states$, where $\overline{\pi}_2$ is defined in Theorem \ref{thm:key_condition} and $c >0$.
\end{assumption}

\begin{theorem}\label{thm:f-small_alpha}
Under Assumptions \ref{ass:lower_bounds},\ref{ass:ergodic} and \ref{ass:optimal_policy-f} with monotone increasing function  $f$ satisfying $f(x) \geq \sqrt{x} $, if $\alpha = \alpha'/|\D|$ satisfying  $\alpha' < C_0\varepsilon_d /\Delta_{\beta}   $ and the following conditions hold:
\begin{align}
  C'_{\mathcal{M}} (\frac{\varepsilon_{\beta}}{f(d^{\D}(\bs_1))} -\sqrt{\varepsilon_{\beta}}) \textbf{Dis}(\overline{\pi}_1, \hatbehavior)(\bs_1)  > (1+c). \label{thm:core_thm:condition}
\end{align}
where $ C'_{\mathcal{M}}$ is a constant independent of $\D$, then (\ref{proof_goal-f}) holds with probability $1-\delta$,

\end{theorem}

\begin{theorem}\label{thm:f-large_alpha}
Assume that  $\alpha > C_0 \max_{\bs}\big(f(\boldsymbol{w}^{\pi/\D})  \cdot \textbf{Dis-CQL}(\pi, \hatbehavior) \cdot \boldsymbol{n}^{-1/2}_{\D} \big)(\bs)$, then  there exists some value $C >1$ such that,  once that  $f(w^{\pi/\D}(\bs))$ is clipped above by $C$, (\ref{proof_goal-f}) holds  with probability $1-\delta$.
\end{theorem}

The analysis for SA-PP still applies to $f$-SA-PP.
The proofs of Theorem \ref{thm:f-key_condition}, \ref{thm:f-small_alpha} and  \ref{thm:f-large_alpha}  can be obtained by making minor modifications to the  proofs of Theorem \ref{thm:key_condition}, \ref{thm:small_alpha_plus} and \ref{thm:large_alpha} with the ratio $\boldsymbol{w}^{\pi/\D}$ replaced by $f(\boldsymbol{w}^{\pi/\D})$. We highlight some key changes of the proof of Theorem \ref{thm:f-small_alpha}:

\textbf{Changes of proof}
According to the proof of Theorem \ref{thm:small_alpha_plus}, we have 
\begin{align}
 \langle \boldsymbol{\freq}_{\D}^{\overline{\pi}_1},C_0\textbf{n}^{-1/2}_{\D,\overline{\pi}_1}\rangle &\geq  \langle \underbrace{\boldsymbol{\freq}_{\D}^{\pi_0},C_0\textbf{n}^{-1/2}_{\D,\pi_0}\rangle}_{\uppercase\expandafter{\romannumeral1}} + \underbrace{\alpha'\frac{1}{|\D|}\langle \boldsymbol{\freq}_{\D}^{\overline{\pi}_1}, f(\frac{\boldsymbol{\freq}_{\D}^{\overline{\pi}_1}}{\boldsymbol{d}^{\D}})\textbf{Dis}(\overline{\pi}_1, \hatbehavior)\rangle - \alpha'\frac{1}{|\D|}\langle \boldsymbol{\freq}_{\D}^{\pi_0}, f(\frac{\boldsymbol{\freq}_{\D}^{\pi_0}}{\boldsymbol{d}^{\D}})\textbf{Dis}(\pi_0, \hatbehavior)\rangle}_{\uppercase\expandafter{\romannumeral2}}
\end{align}

For the second term, we have
\begin{align}
\uppercase\expandafter{\romannumeral2} 
\geq - \alpha' \Delta_{\beta} \frac{1}{|\D|\cdot f(d^{\D}(\bs_1))},
\end{align}
if $f(x) \geq x$, we have 
\begin{align}
\uppercase\expandafter{\romannumeral2} 
\geq - \alpha' \Delta_{\beta} \frac{1}{|\D|\cdot d^{\D}(\bs_1)} = - \alpha' \Delta_{\beta}/n_{\D}(\bs_1),
\end{align}
then the rest proof of Theorem \ref{thm:small_alpha_plus} still holds for Theorem \ref{thm:f-small_alpha}.

if $\sqrt{x} \leq f(x) < x$, we have 
\begin{align}
\uppercase\expandafter{\romannumeral2} 
\geq - \alpha' \Delta_{\beta} \frac{1}{f(|\D|)\cdot f(d^{\D}(\bs_1))} \geq - \alpha' \Delta_{\beta} \frac{1}{\sqrt{n_{\D}(\bs_1)}},
\end{align}
then the rest proof of Theorem \ref{thm:small_alpha_plus} still holds for Theorem \ref{thm:f-small_alpha}.

\section{Pseudo-code of SA-CQL}\label{app:pseudo_code}
 Algorithm \ref{alg:practical_alg} is the pseudo-code of SA-CQL.
\begin{algorithm}[t]
\small
\caption{State-Aware Conservative Q-Learning, differences with  \cite{kumar2020conservative} are colored}
\label{alg:practical_alg}
\begin{algorithmic}[1]
    \STATE Initialize Q-function, $Q_\theta$,  policy $\pi_\phi$, density estimator $\nu_{\theta_1}$ and $\zeta_{\theta_2}$,
    \STATE \textcolor{red}{Pre-train $\nu_{\theta_1}$, $\zeta_{\theta_2}$ and corresponding $\omega^{\pi_t/\D}$ by solving the objectives of Equation (\ref{equ:dualDICE})  with all data using $G_{pre}$ gradient steps}
    \FOR{step $t$ in \{1, \dots, N\}}
    \STATE \textcolor{red}{Obtain $\nu_{\theta_1}$, $\zeta_{\theta_2}$ and corresponding $\omega^{\pi_t/\D}$ by solving the objectives of Equation (\ref{equ:dualDICE}) with  $G_{\zeta}$ gradient steps}
        \STATE Train the Q-function using $G_Q$ gradient steps on objective from Equation~(\ref{eqn:practical_objective}) \\
        \mbox{$\theta_t := \theta_{t-1} - \eta_Q \nabla_\theta \textcolor{red}{\text{SA-CQL}(\theta)}$}\\
        \STATE  Improve policy $\pi_\phi$ via $G_\pi$ gradient steps on $\phi$ with SAC-style entropy regularization:\\
        \mbox{$\phi_{t} := \phi_{t-1} + \eta_\pi \mathbb{E}_{\bs \sim \mathcal{D}, \ba \sim \pi_\phi(\cdot|\bs)}[Q_\theta(\bs, \ba)\! -\! \log \pi_\phi(\ba|\bs)] $}
    \ENDFOR
\end{algorithmic}
\end{algorithm}

It is noted that, empirically,  \textbf{it is unnecessary  to use all data to estimate the density ratio at ``every" step}. A high-quality  estimator ($\nu$ and $\zeta$) is pre-trained at the beginning of the training process, which only brings in few extra gradient steps.  During the policy training process, the pre-trained estimator and the policy will be updated together 
using the same batch and comparable gradient steps,  since  the policy is slowly changing. Overall, extra cost is  favorable   thanks to the delicate design of DualDICE  and thus SA-CQL is  much cheaper than many ensemble-based methods, especially   a strong baseline EDAC~\cite{an2021uncertainty} which requires  $10-50$ ensembles.

\section{Implementation Details of SA-CQL}\label{app:imp_detail}
The experiments are conducted on an Intel(R) Xeon(R) Gold 6134 processor based Ubuntu 18.04.6 LTS Server, which consists of one processor of 16 cores, running at 3.20GHz with 32KB of L1, 1024KB of L2, 25344KB of L3 cache, and 128GB of memory and 1 Quadro RTX 5000 GPU. The MuJoCo Gym datasets we used in our experiments are v2 versions, which fixed some bugs as reported here \cite{d4rl}. Our codes are implemented with Python 3.6 and PyTorch.  
 The results for BCQ, BEAR, CQL and TD3plusBC are from our own re-implementation  based on open-source library d3rlpy \cite{seno2021d3rlpy} 
following MIT license, the hyper-parameters settings for these algorithms also follow\cite{seno2021d3rlpy}  as well.  The results for UWAC, F-BRC and EDAC are taken from the author-provided open-source  and follow their original hyper-parametrs settings, respectively. The results for REM are taken from 
the author-provided open-source of \cite{agarwal2020optimistic} and follow their original hyper-parametrs settings.

For SA-CQL algorithm, we use  the default CQL/discrete CQL implementation of  \cite{seno2021d3rlpy} to suit our environments,
and based on which we implement  SA-CQL.
For DualDICE estimators, we resort to the official  \cite{nachum2019dualdice} to ease our implementation, for both continuous and discrete control setting.
Specially, for discrete control setting, the outputs of the encoder serve as the inputs of  $\mu$ and $\zeta$
networks.
Some crutial hyperparameters for SA-CQL are shown in Table \ref{app:hyper_parameters-1}-\ref{app:hyper_parameters-1-1}.

We run each algorithm for one million training steps and report the normalized average return of each policy. The normalized average return is computed using the D4RL built-in \textit{env.get\_normalized\_score(returns)} function where the return is the accumulated un-discounted rewards of an episode.  Each algorithm is evaluated with three different seeds and the performance of each policy is evaluated for $10$ episodes. 

To realize $f(x) = b_1\cdot (\max_x \log x - \min_x \log x(\log x - \min_x \log x)+b_0$, we  use the minimum and maximum of a mini batch to approximate $\min_{x}\log x$ and $\max_{x}\log x$.
In particular, to make the ratios' estimates more stable, some pre-training is conducted:
we pre-train a policy using CQL for $20000$ steps, then  fix the learned  $Q_{\theta}$, $\pi_{\phi}$ and separately train the DualDICE estimator $\nu_{\theta_1}, \zeta_{\theta_2}$ for $G_{pre}=100000$ steps.
After  pre-training, we set the learned  $Q_{\theta}$, $\pi_{\phi}$, $\nu_{\theta_1}$ and $\zeta_{\theta_2}$   as initials and follow  Step 3-6 of Algorithm \ref{alg:practical_alg} to keep training. The gradient steps $G_{\zeta}, G_{Q}, G_{\pi}$  are all $1$.

\section{Discussion about $b_0$ and $b_1$}
The hyper-parameters settings for $b_0$ and $b_1$ are presented  in Table \ref{app:hyper_parameters-2}. We set $b_1$ larger on ``expert" and ``medium-expert" datasets since we conjecture that their data distributions are narrower than other datasets which may need bigger conservativeness. It is also observed that the estimated ratios $\boldsymbol{w}^{\pi/\D}$  for a random $\pi$ on these two datasets are  remarkably larger  than those on the other datasets within the same environment, which reveals that  the ratios are also informative for hyper-parameters tuning.

Since we set the upper bound of  the  state distribution ratios $b_1$ as $5$ for some datasets, which may make SA-CQL more conservative than CQL due to the composite effect of $b_1 \cdot \alpha$. 
To guarantee that CQL cannot be improved only by changing conservative weight, or say, SA-CQL outperforms  due to state-aware pessimism rather than tricky hyper-parameter setting, we conduct another ablation study  to compare SA-CQL with CQL using different $\alpha$. The results  in  Table  \ref{table:conservative} show that
SA-CQL  still outperforms the best CQL baseline on almost all datasets and remarkably outperforms it on half of the datasets. This reveals  state-aware pessimism is the necessity for performance improvement.

We additionally conduct experiments by setting $b_0 = 0.5$ for halfcheetah and hopper, so that all the datasets share the same $b_0$ and the results are shown in Table \ref{app:table:union_b0}, we can see that there are only some slight drops for some datasets, and the average score is $78.4$ which still outperforms the other baselines, showing that our method is robust for the hyper-parameters.

\begin{table}
    \centering
    \caption{Hyperparameters for SA-CQL on continuous control setting.}
    \scalebox{0.9}{
    \begin{tabular}{l|r}
        \hline
        \textbf{Hyperparameter} & $ \textbf{Value} $    \\
        \hline
         Critic learning rate & $3e-4$ \\
         Actor learning rate & $1e-4$\\
         Fixed conservative weight  & 5  \\
         Mini-batch size & 256 \\
         Action samples number & 10 \\
         \midrule
         $\zeta$ learning rate & 1e-4 \\
         $\mu$ learning rate & 1e-4 \\
         average samples number  & 1. \\
         $\nu$ hidden units & [256,  256]\\
         $\zeta$ hidden units & [256, 256]\\
         \hline
    \end{tabular}
    }
    
    \label{app:hyper_parameters-1}
\end{table}

\begin{table}
    \centering
    \caption{Hyperparameters for SA-CQL on discrete control setting.}
    \scalebox{0.9}{
    \begin{tabular}{l|r}
        \hline
        \textbf{Hyperparameter} & $ \textbf{Value} $    \\
        \hline
         Critic learning rate & $6.25e-5$ \\
         Fixed conservative weight  & 1  \\
         Mini-batch size & 64 \\
         \midrule
         $\zeta$ learning rate & 1e-3 \\
         $\mu$ learning rate & 1e-4 \\
         average samples number  & 1. \\
         $\nu$ hidden units & [256,  256]\\
         $\zeta$ hidden units & [256, 256]\\
         \hline
    \end{tabular}
    }
    
    \label{app:hyper_parameters-1-1}
\end{table}

\begin{table}[!htbp]
    \centering
    \caption{Hyperparameters choices of  $b_0$ and $b_1$ for different offline datasets}
    \scalebox{0.9}{
    \begin{tabular}{l|r|r}
    \hline
        \textbf{Dataset} & $b_0$ & $b_1$    \\
        \hline
         halfcheetah-medium & 0. & 1. \\
         halfcheetah-medium-replay & 0.&  1.\\
         halfcheetah-full-replay & 0. & 1. \\
         halfcheetah-expert & 0.& 5. \\
         halfcheetah-medium-expert & 0. & 5.\\
         halfcheetah-random & 0. & 1.\\
         \midrule
         hopper-medium & 0.&1.\\
         hopper-medium-replay & 0. &1.\\
         hopper-full-replay &0. & 1.\\
         hopper-expert & 0. &  5.\\
         hopper-medium-expert & 0. & 5.\\
         hopper-random & 0. & 1.\\
         \midrule
         walker2d-medium &0.5 & 1.\\
         walker2d-medium-replay & 0.5 &1.\\
         walker2d-full-replay  &0.5 & 1. \\
         walker2d-expert  &0.5 & 5.\\
         walker2d-medium-expert  &0.5 & 5.\\
         walker2d-random  &0.5 & 1.\\
         \hline
         Pong & 0.5 & 2. \\
         Qbert & 0.5 &  2.\\
         Seaquest & 0.5& 2. \\
         Breakout & 0.5 & 2. \\
         \hline
    \end{tabular}
    }
    
    \label{app:hyper_parameters-2}
\end{table}

    

\begin{table}[!htbp]\caption{Comparison of SA-CQL to CQL with different conservative weights}\label{table:conservative}
\centering
\scalebox{0.75}{
\begin{tabular}{l|cccc|cccc|cccc|}
\hline
$\alpha$ & \multicolumn{4}{c|}{halfcheetah}                                                                                                             & \multicolumn{4}{c|}{hopper}                                                                                                           & \multicolumn{4}{c|}{walker}                                                                                                                    \\
         & m                                 & m-r                               & f-r                               & r                                & m                                 & m-r                       & f-r                              & r                                  & m                                 & m-r                               & f-r                                & r                                 \\ \hline
2.5      & 57.                               & 51.6                              & 82.6                              & 27.1                             & 71.3                              & \textbf{102}              & 85.7                             & 6.3                                & 0.                                & 52.0                              & 101.0                              & 1.9                               \\
5        & 52.5                              & 49.3                              & 80.5                              & 26.2                             & 74.1                              & 90.3                      & 107.3                            & 12.1                               & 85.4                              & 82.9                              & 97.7                               & 0.                                \\
10       & 49.4                              & 47.56                             & 78.6                              & 18.6                             & 73.4                              & 97.2                      & 103.                             & 8.0                                & 83.4                              & 86.2                              & 94.9                               & 0.                                \\
25       & 46.4                              & 45.3                              & 76.8                              & 12.8                             & 61.8                              & 98.3                      & 100.8                            & 7.5                                & 82                                & 72.6                              & 92.5                               & 3.9                               \\ \hline
SA-CQL   & \multicolumn{1}{l}{\textbf{58.1}} & \multicolumn{1}{l}{\textbf{55.1}} & \multicolumn{1}{l}{\textbf{83.1}} & \multicolumn{1}{l|}{\textbf{31}} & \multicolumn{1}{l}{\textbf{86.3}} & \multicolumn{1}{l}{100.1} & \multicolumn{1}{l}{\textbf{108}} & \multicolumn{1}{l|}{\textbf{17.7}} & \multicolumn{1}{l}{\textbf{87.7}} & \multicolumn{1}{l}{\textbf{90.1}} & \multicolumn{1}{l}{\textbf{102.3}} & \multicolumn{1}{l|}{\textbf{4.1}} \\ \hline
\end{tabular}
}
\end{table}

\begin{table}
    \centering
     \caption{SA-CQL for $b_0=0.5$ for halfcheetah and hopper}
\begin{tabular}{l|r|}
\hline
\textbf{Dataset}          & $b_0$           \\ \hline
halfcheetah-medium        & 54.2 $\pm$ 0.5  \\
halfcheetah-medium-replay & 50.6 $\pm$ 0.8  \\
halfcheetah-full-replay   & 82.4 $\pm$ 0.3  \\
halfcheetah-expert        & 96.8 $\pm$ 2.1  \\
halfcheetah-medium-expert & 85.8 $\pm$ 3.   \\
halfcheetah-random        & 27. $\pm$ 0.3   \\ \hline
hopper-medium             & 76.8 $\pm$ 2.3  \\
hopper-medium-replay      & 101.5 $\pm$ 0.3 \\
hopper-full-replay        & 107.2 $\pm$ 0.3 \\
hopper-expert             & 112.5 $\pm$ 0.2 \\
hopper-medium-expert      & 104. $\pm$ 0.5  \\
hopper-random             & 8.1 $\pm$ 0.2              \\ \hline
\end{tabular}

    \label{app:table:union_b0}
\end{table}

\bibliographystyle{unsrt}  

\end{document}